\documentclass{article}

\usepackage{jfrExamplee}
\usepackage{graphicx}
\usepackage{apalike}
\usepackage{setspace}

\usepackage{amsmath,amsfonts,amssymb}
\usepackage{tikz,tkz-euclide}
\usetikzlibrary{arrows,calc,shapes,positioning}
\tikzset{>=latex}
\usepackage{color,soul}
\usepackage{subcaption}
\usepackage{pgfplots}
\usepackage{multirow}
\usepackage[colorinlistoftodos]{todonotes}
\usepackage{changes}
\usepackage{svg}
\graphicspath{ {images/} }
\usepackage{colortbl}

\usepackage[ruled, noend]{algorithm2e}
\usepackage{multirow}
\usepackage{cite}
\usepackage{cleveref}

\usepackage{adjustbox}
\usepackage{makecell}
\usepackage{tabularx}

\usepackage{lscape}

\newcommand\Tstrut{\rule{0pt}{2.6ex}}

\title{\LARGE \bf
 GPGM-SLAM: a Robust SLAM System for Unstructured Planetary Environments with Gaussian Process Gradient Maps
}

\author{
Riccardo Giubilato \\
Institute of Robotics and Mechatronics\\
German Aerospace Center (DLR)\\
\texttt{riccardo.giubilato@dlr.de} \\
\And
Cedric Le Gentil \\
Centre for Autonomous Systems \\
University of Technology Sydney \\
\texttt{cedric.legentil@student.uts.edu.au} \\
\AND
Mallikarjuna Vayugundla \\
Institute of Robotics and Mechatronics\\
German Aerospace Center (DLR)\\
\texttt{mallikarjuna.vayugundla@dlr.de} \\
\And
Martin~J.~Schuster \\
Institute of Robotics and Mechatronics\\
German Aerospace Center (DLR)\\
\texttt{Martin.Schuster@dlr.de} \\
\And
Teresa Vidal-Calleja \\
Centre for Autonomous Systems \\
University of Technology Sydney \\
\texttt{teresa.vidalcalleja@uts.edu.au} \\
\And
Rudolph Triebel \\
Institute of Robotics and Mechatronics\\
German Aerospace Center (DLR)\\
\texttt{Rudolph.Triebel@dlr.de} \\
}

\begin{document}

\maketitle

\def\nbpoints{N}
\def\i{i}
\def\m{m}
\def\nbsubmaps{M}
\newcommand\submap[1]{\mathcal{S}^{#1}}
\newcommand\point[2]{\mathbf{p}_{#2}^{#1}}
\newcommand\x[2]{x_{#2}^{#1}}
\newcommand\y[2]{y_{#2}^{#1}}
\newcommand\z[2]{z_{#2}^{#1}}
\newcommand\xc[1]{x^{#1}}
\newcommand\yc[1]{y^{#1}}
\newcommand\zc[1]{z^{#1}}
\def\zmat{\mathbf{Z}}
\newcommand\inputdomain[1]{\mathbf{x}^{#1}}
\newcommand\inputdomaini[2]{\mathbf{x}^{#1}_{#2}}
\def\inputdomainmat{\mathbf{X}}
\newcommand\elevationfunction[1]{f({#1})}

\newcommand{\todoR}[1]{{\color{blue!80!green}\textbf{@Riccardo}: [#1]}}
\newcommand{\todoA}[1]{{\color{green!80!blue!50!black}\textbf{@Arjun}: [#1]}}
\newcommand{\todoC}[1]{{\color{red!80!blue!50!black}\textbf{@Cedric}: [#1]}}
\newcommand{\todoAll}[1]{{\color{yellow!80!red!50!black}\textbf{@Todo}: [#1]}}

\begin{abstract}
	Simultaneous Localization and Mapping (SLAM) techniques play a key role towards long-term autonomy of mobile robots due to the ability to correct localization errors and produce consistent maps of an environment over time.
	Contrarily to urban or man-made environments, where the presence of unique objects and structures offer unique cues for localization, the apperance of unstructured natural environments is often ambiguous and self-similar, hindering the performances of loop closure detection.
	In this paper, we present an approach to improve the robustness of place recognition in the context of a submap-based stereo SLAM based on Gaussian Process Gradient Maps (GPGMaps). 
	GPGMaps embed a continuous representation of the gradients of the local terrain elevation by means of Gaussian Process regression and Structured Kernel Interpolation, given solely noisy elevation measurements.
	We leverage the image-like structure of GPGMaps to detect loop closures using traditional visual features and Bag of Words. GPGMap matching is performed as an SE(2) alignment to establish loop closure constraints within a pose graph. 
	We evaluate the proposed pipeline on a variety of datasets recorded on Mt.\ Etna, Sicily and in the Morocco desert, respectively Moon- and Mars-like environments, and we compare the localization performances with state-of-the-art approaches for visual SLAM and visual loop closure detection.
\end{abstract}

\section{Introduction}
Within the last decades, the field of mobile robotics underwent a significant technological leap in terms of enhanced
mobility, sensory perception and planning of actions.
The resulting increased capabilities are crucial, within our modern society, as they provide tools for disaster
mitigation \cite{klamt2018supervised}, search and rescue operations \cite{de2013eu,delmerico2019current} and
exploration of hazardous and unreachable environments, such as in the case of exploration of planetary bodies
\cite{wedler2021german}.
In many occasions, the robotic assets need to operate autonomously, without relying on the intervention of human
operators.
This is the case of robotics for the exploration of Mars or the Earth's moon, where communication delays limit the
possibility of manual intervention \cite{heverly2013traverse}.
On Earth, the scenario of subterranean exploration imposes similar constraints,
due to the possibility of communication dropouts \cite{otsu2020supervised}.

A key skill towards prolonged robotic autonomy is the ability to localize in previously unknown environments without
the need of global positioning systems.
Simultaneous Localization and Mapping (SLAM) \cite{durrant2006simultaneous} solves this problem by utilizing sensory
inputs to concurrently build a map, or a representation of the environment, and localize an observer with respect to it.
Visual SLAM techniques exploit cameras for ego-motion estimation, mapping and place recognition and, to this day, are
mature enough to find commercial application in the domains of autonomous driving \cite{Cadena2016} or augmented
reality (AR) \cite{li2017monocular}.

\begin{figure}[t]
\includegraphics[width=\linewidth]{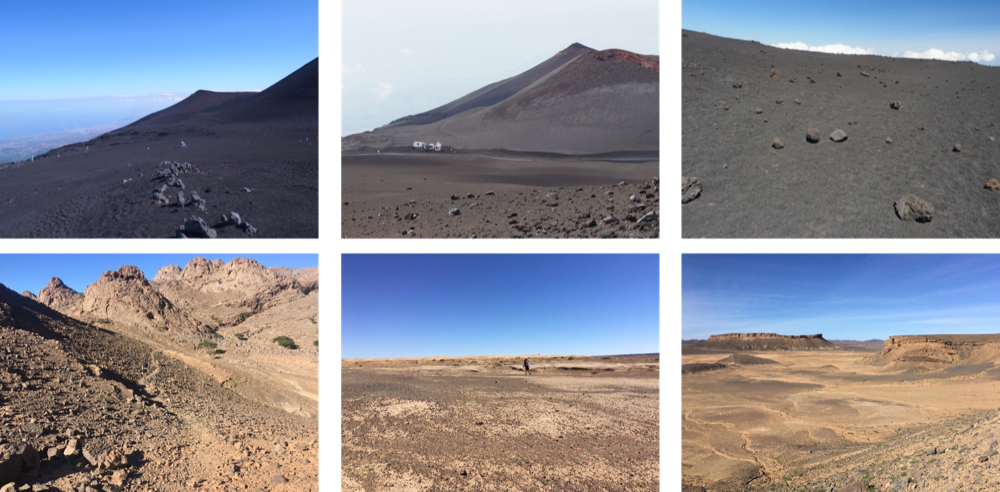}
	\caption{Impressions of the Moon-like landscape on the volcano Mt.\ Etna (top) and of the Mars-like landscape in the Moroccan desert (bottom).}
	\label{fig:intro:landscapes}
\end{figure}

Traditional Visual SLAM approaches \cite{Mur-Artal2017,qin2018vins}, based on detection and
tracking of visual features, are proven to be accurate and robust against failures and drift in the pose estimation
over time, thanks to well-developed loop closure detection techniques \cite{Galvez-Lopez2012,cummins2008fab}, which
allow to establish global optimization problems to correct trajectory and map by enforcing pose constraints while
revising places.
However, the typical scenarios where this family of algorithms is developed and tested comprise often
man-made or, in general, feature-rich environments \cite{geiger2013vision,sturm2012benchmark}.
As an example, the task of visual place recognition from the point of view of a moving car is subject to strong prior
constraints on the camera perspective: the car would likely revisit a place from the same direction or the
opposite, as its motion is constrained by the lanes on the road.

Contrarily, visual place recognition in unstructured natural environments is challenging due to aliasing and lack of unique visual or structural features that suggest
unambiguously a previously visited location \cite{giubilato2020relocalization}.
This is the case in the planetary exploration scenario, where unique landmarks are scarce and hardly recognizable due
to the environment appearance as well as the perceptive behaviors of mobile robots performing autonomous navigation
in such context.
The necessity to use visual inputs for the purpose of traversability estimation and hazard avoidance results in
arbitrary viewpoints for cameras, usually mounted on pan-tilt units, hindering the likelihood of collecting images of
the same landscapes from repeatable positions and view directions.
This is further aggravated by the fact that robots do not necessarily traverse the same paths as there are typically no designated roads or pathways on such rough terrains.

This paper builds on our previous work~\cite{gentil2020gaussian} where we introduced the concept of Gaussian
Process Gradient Maps (GPGMaps): local representations of the environment obtained by fitting an elevation model on
submaps~\cite{schuster2018distributed} through Gaussian Process regression and applying linear operators to compute
the spatial derivatives.
Thus, instead of relying on the visual similarity of camera images, the similarity of the
terrain elevation is leveraged to both, select candidate matches between GPGMaps as well as establishing loop closures.
Through Gaussian Process regression, GPGMaps represent the environment in a continuous manner, robust to
measurement noise, allowing to detect the similarity of the terrain regardless of the direction of travel and the
distance of the viewpoints.
We exploit GPGMaps to build a graph-based SLAM architecture targeted at stereo vision-based perception systems for
robots operating in challenging natural and unstructured environments.

Thanks to the involvement in the Helmholtz alliance ROBEX "Robotic Exploration of EXtreme Environments"
\cite{Wedler2017FirstRO} in 2012-2017 and the project ARCHES "Autonomous Robotic Networks to Help Modern Societies"
\cite{schuster2020arches},
2018-2022<, the DLR (German Aerospace Center) Institute of Robotics and Mechatronics is conducting specific research on
enabling autonomy for an team of heterogeneous robots which comprises an UAV, ARDEA \cite{lutz20ardea}, and a
planetary-like rover, the Lightweight Robotic Unit (LRU)~\cite{Schuster2017TowardsAP}.
Within the final demonstration mission of project ROBEX, a variety of datasets has been recorded with accurate
differential GPS ground truth, where the LRU robots perform different tasks, e.g. waypoint navigation, autonomous
exploration, and long-range traversing~\cite{vayugundla2018datasets}.
The mission took place on Mt. Etna, Sicily, a volcanic environment designated as a Moon-analogue site and
especially challenging for the purpose of localization for the reasons previously introduced.

\begin{figure}[t]
	\centering
	\begin{subfigure}{0.48\textwidth}
		\includegraphics[width=\linewidth]{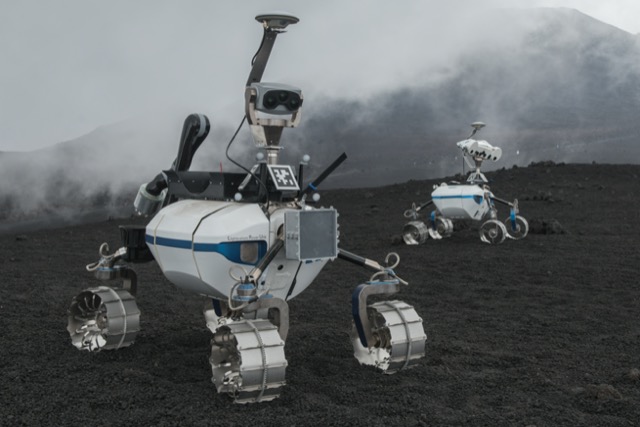}
	\end{subfigure}\hfill
  \begin{subfigure}{0.48\textwidth}
		\includegraphics[width=\linewidth]{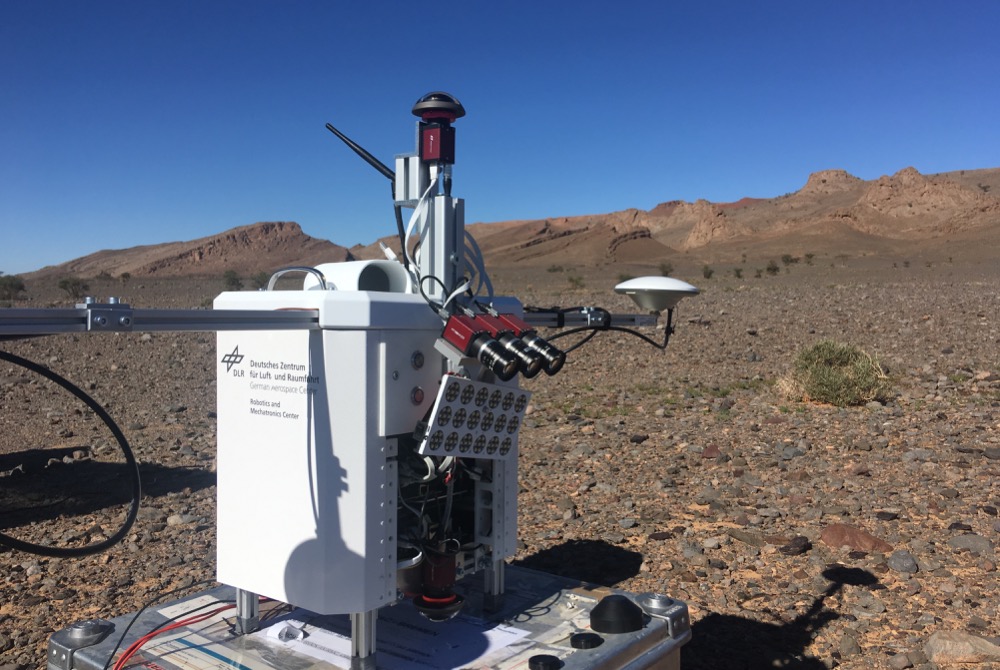}
  \end{subfigure}
	\caption{Pictures of the Lightweight Rover Unit (LRU) rovers on Mt. Etna (left) and the Sensor Unit for Planetary Exploration Rovers (SUPER) handheld system in the Morocco desert (right).}
	\label{fig:systems}
\end{figure}

In the context of collecting datasets in remote planetary-like environments, in 2018 the DLR Institute of Robotics
and Mechatronics recorded the Morocco-Acquired Dataset of Mars-Analogue eXploration (MADMAX) \cite{lukas2021MADMAX} on a variety of Mars-analogue sites in Morocco.
This dataset specifically addresses the purpose of benchmarking localization algorithms in challenging Mars-like
scenarios and was recorded with a perception system, the Sensor Unit for Planetary Exploration Rovers (SUPER),
analogous to the one of the LRU rovers and the ARDEA UAV.
Figure~\ref{fig:intro:landscapes} shows a variety of views of the Etna and Morocco sites respectively on the top and bottom rows, highlighting the similarity of both locations to lunar and martian scenes.
Figure~\ref{fig:systems} gives instead an impression of the LRU rover and the SUPER system.

By evaluating the proposed SLAM system on the Mt.\ Etna and Morocco datasets, we analyze the performance of our
approach in unstructured environments, which closely represent the challenges in localization and mapping brought by planetary-like scenarios where no prior assumptions can be made.
We furthermore compare the localization performances of GPGM-SLAM with state of the art visual SLAM systems, ORB-SLAM2 \cite{Mur-Artal2017} and VINS-Mono \cite{qin2018vins}, as well as its pure place recognition capabilities with the visual loop closure detector iBoW-LCD \cite{garcia2018ibow}.
We therefore highlight the specific problems that hinder the chances of establishing loop closures and report our lessons learned for the researchers working on SLAM systems on similar field robotics applications.

More specifically, this paper presents the following contributions going beyond our previous works on GPGMaps:

\begin{itemize}
\item we improve on \cite{gentil2020gaussian} by implementing an efficient GP regression method based on Structured Kernel Interpolation (SKI) \cite{wilson2015kernel} to infer the gradient of the terrain elevation using solely noisy elevation measurements. SKI allows us to sample the GP output on a denser domain and reduce the computational complexity from $\mathcal{O}(n^3)$ to $\mathcal{O}(n)$. Furthermore, we propose a variation of the SKI method, which permits to generate gradients with only elevation as input.
\item we present a full SLAM system based on GPGMaps with loop closing capabilities showing unprecedented
localization performances on completely unstructured scenarios.
\item we perform extensive testing of our pipeline on challenging datasets recorded during field tests on Mt.\ Etna, Sicily and in the Morocco desert, which are respectively designated as Moon- and Mars-analogue environments. We demonstrate higher localization performances compared to our current SLAM system, targeted at teams of heterogeneous robots \cite{schuster2018distributed}, and compared to the state of the art visual SLAM systems ORB-SLAM2 and VINS-MONO.
\item we analyze in detail the properties of structure- and visual-based loop closure detection in the context of unconstrained camera trajectories in natural environments. We then demonstrate how our approach leads to higher chances of detecting previously observed terrain structures compared to pure visual similarity, also when revisiting places from opposite travel directions.
\end{itemize}

This paper is structured as follows: In Section~\ref{sec:related_works} we give an overview of related works about SLAM in outdoor unstructured environments, in Section~\ref{sec:SLAM_arch} we introduce the components of our SLAM system and the integration of GPGMap matching as an alternative to our baseline configuration. In Section~\ref{sec:GPGM} we introduce the concept of GPGMap and in Section~\ref{sec:LC_detection} we explain how GPGMaps are used in the context of detecting loop closures. Section~\ref{sec:bow_eval} presents a preliminar assessment of the precision of selecting candidate GPGMap matches using the BoW paradigm. Section~\ref{sec:eval} contains the evaluation of GPGM-SLAM on a variety of sequences in planetay analogue environments and is followed by a discussion on the different behaviour of structure- and visual-based loop closure detection.
Finally, we draw the conclusions and potential developments of GPGM-SLAM in Section~\ref{sec:conclusions}.

\section{Related Works on Outdoor SLAM}
\label{sec:related_works}
Visual localization in challenging outdoor scenarios has been successfully performed for many decades now.
A notable example is the implementation of Visual Odometry (VO) onboard the Mars Exploration Rovers (MER) Spirit and Opportunity \cite{olson2003rover} to decrease localization drift accumulated from the wheel slippage on soft grounds.
This allowed the rovers to travel on longer traverses for many years without the need of global localization \cite{maimone2007two}.
Although VO techniques have been proven to be a robust manner to estimate a vehicle's ego motion, the ever present drift, albeit low, is never compensated.
To this end, Simultaneous Localization and Mapping (SLAM) techniques \cite{durrant2006simultaneous} have been developed to solve both the localization and mapping problems concurrently, reducing localization drift thanks to the ability of establishing loop closures when revisiting known places.
The earliest formulations of the probabilistic SLAM problem leveraged Kalman Filters (EKF) \cite{bailey2006consistency} and particle filters \cite{montemerlo2003fastslam}, the first limited by the growing computational complexity with the increasing number of landmarks and the second limited by the memory consumption proportional to the number of particles.
A turning point was reached with graph-based formulations \cite{thrun2006graph,grisetti2010tutorial} and smoothing approaches \cite{kaess2008isam,kaess2012isam2} where the sparsity of the SLAM problem was leveraged to increase computational efficiency.

Visual SLAM is traditionally performed through detection and tracking of visual features \cite{bay2006surf,rublee2011orb,lowe2004distinctive}.
The task of recognizing previously visited places relies on detecting the occurrence of already observed features, mostly by means of aggregation techniques such as Bag of Words \cite{GalvezTRO12} or Vectors of Locally Aggregated Descriptors (VLAD) \cite{arandjelovic2013all}.
In recent years, visual SLAM systems, such as ORB-SLAM2 \cite{Mur-Artal2017} or LSD-SLAM \cite{engel2014lsd}, reached a maturity such that they can successfully be deployed in mobile robots or devices to perform mapping and localization in feature-rich environments.
To this day, however, modern visual SLAM systems have yet to demonstrate successful long term operations in unstructured and homogeneous scenarios, while excelling in the fields of augmented reality and autonomous driving \cite{bresson2017simultaneous,cadena2016past}.
The self similarity and lack of diverse visual features, in fact, challenges the ability of recognizing already visited places and therefore the possibility to compensate localization drift.
This is aggravated by the strict dependency of visual place recognition on the repeatability of viewpoints, which is not guaranteed for the case of unconstrained robot navigation in outdoor scenes.
To this end, earlier approaches, targeting the specific application scenario of exploring planetary-like environments, leveraged visual saliency to segment repeatable regions in the image and track them through visual features \cite{bajpai2016planetary}. Although compensating for the self-similarity of the environment and minimizing the computational load, the possibility of detecting loop closures is still dependent on the repetition of viewpoints.
The presence of pan-tilt mechanisms on planetary rovers allows to plan the camera viewpoints in order to maximize the accuracy of VO \cite{otsu2017look} and potentially facilitating the detection of loop closures, the success of which is however dependent on the localization uncertainty and can become hard to tackle.
A different path towards visual place recognition belongs to the SeqSLAM approach \cite{milford2012seqslam,siam2017fast}, proven to work with some adaptations also on very challenging unstructured and self-similar scenarios \cite{grixa2018appearance}.
This approach, however, consists in matching continuous sequences of images, which poses again strong constraints on the sequence of viewpoints and is more targeted to path following or homing tasks.

The utilization of range sensors in the context of SLAM helps to overcome the limitations of traditional visual sensing.
Earlier approaches leveraged plane scanning LiDARs mounted on pan-tilt units to produce scans with 360 degree coverage, using the scans for either obstacle detection \cite{schafer20083d} or trajectory planning and localization \cite{rekleitis2009autonomous} in outdoor environments.
Following the recent democratization of LiDAR sensors, many odometry and SLAM systems have been developed using 3D LiDARs as the main sensing source \cite{zhang2014loam,shan2018lego,legentil2021in2laama} and demonstrated to work reliably and accurately in urban landscapes where the surrounding structures (e.g., buildings, cars, and trees) well constrain rigid registration algorithms such as Iterative Closest Point (ICP).
Even though modern LiDAR odometry methods achieve high accuracy in pose estimation on long travels, the ever growing localization drift must be compensated by establishing loop closures.
Many approaches for loop closure detection based on LiDAR scans have been developed recently and follow different strategies.
Similarly to visual methods, the detection of 3D features, such as SHOT \cite{salti2014shot} or FPFH \cite{rusu2009fast} and their variations, allow to perform place recognition by matching keypoints from point clouds in moderately unstructured scenarios.
In \cite{guo2019local}, SHOT descriptors augmented by LiDAR intensity are used in a probabilistic voting scheme to find candidate point cloud matches from a database.
Intensity of LiDAR measurements is used also in \cite{cop2018delight}, to produce spatial histograms as global point cloud descriptors.
Other approaches employ projection of ranges or intensity from point clouds to perform feature extraction on images. NARF (Normal Aligned Radial Features) are used in \cite{6094638} in a Bag of Word scheme to find candidate matches from urban and indoor scenes, and discard ambiguous point clouds through a self similarity test.
Other approaches rely on global descriptors of LiDAR scans, such as ScanContext \cite{kim2018scan}, or on segmentation and accumulation of distinct point clusters, such as SegMap \cite{Dube-RSS-18,dube2020segmap}.
The dependency on specific assumptions about the target environment, such as the presence of evident structures that can be easily segmented either by removing a traversable ground or by matching primitive shapes \cite{pierzchala2018mapping}, makes these approaches unsuitable for generic natural and planetary-like landscapes.
Furthermore, to this day no mechanical or, in general, 3D LiDAR sensor has been deployed on an autonomous rover mission for planetary exploration, with the exception of NASA's Ingenuity helicopter which carried an 1D LiDAR altimeter for the purpose of altitude estimation. Therefore, in our work we will leverage structural information but in the context of stereo vision.

Among SLAM systems that specifically address the target application of rovers in planetary-like scenarios, \cite{hidalgo2018adaptive} implements a SLAM system where an odometry model is inferred through the usage of Gaussian Process regression to complement a visual pipeline where the extracted keyframes are added to a graph in an adaptive manner, as to maximize the information gain and minimize the computational load.
The overall system has been successfully tested on analogue planetary terrains, where however the authors highlight the challenging nature of the acquired images, difficult to use for loop closure detection due to the strong perceptual aliasing.
The authors of \cite{geromichalos2020slam} propose a SLAM system based on a particle filter that includes global localization constraints.
2.5D local maps are built from stereo data and matched to the current map using a variant of ICP, whose scores are used to resample particles.
Global localization is performed by performing template matching between local and global elevation images.
The method is interesting and it shares with our system the idea of using local elevation and registration techniques from the domain of image matching to perform localization.
However, the usage of plain elevation is insufficient to provide enough visual details for matching and the transformation of local point clouds to 2.5D maps loses the 3D information that is delivered by the original point clouds.

\section{The SLAM Architecture}
\label{sec:SLAM_arch}
\begin{figure}[t]
	\centering
	{
	\fontfamily{lmss}\selectfont
	\includesvg[width=\linewidth]{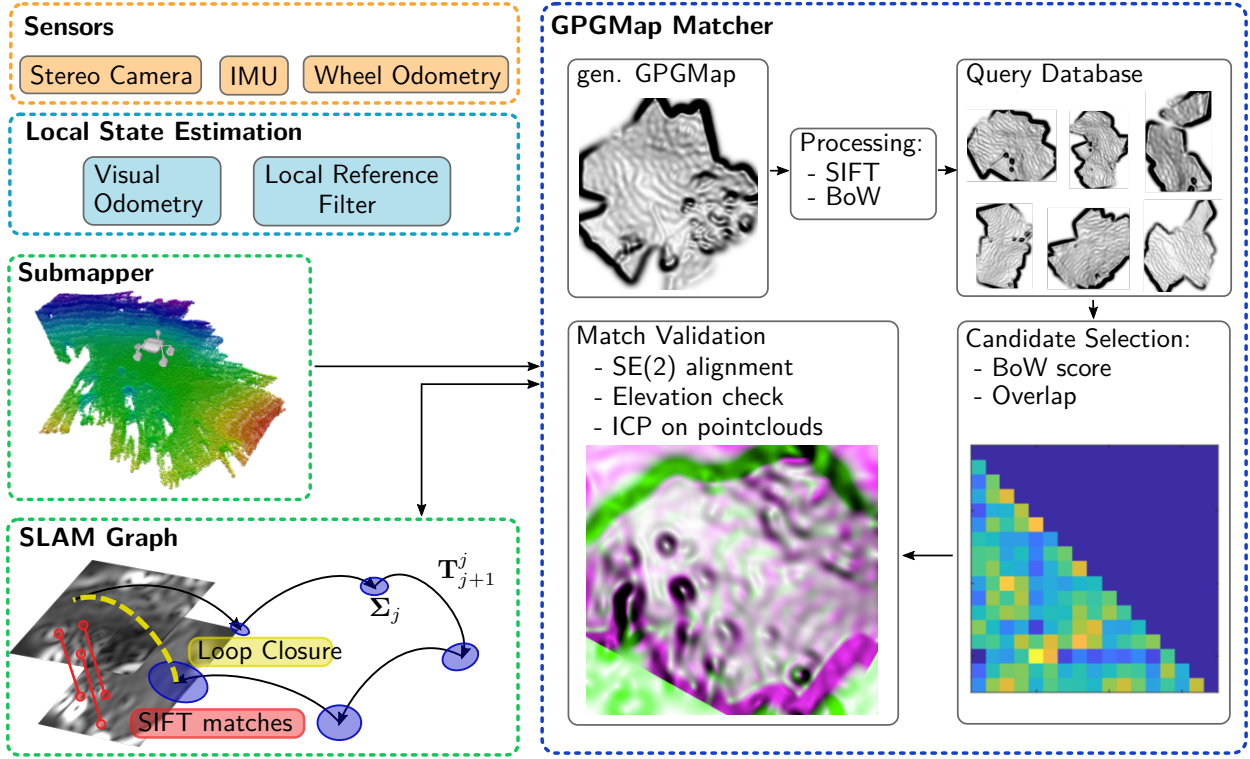}
	}
	\caption{Overview of the GPGM-SLAM pipeline: (a) GPGMaps are generated using 3D submaps from aggregated stereo clouds. Visual features, computed on the gradient images, are used to create BoW vectors and detect loop closures from a database. (b) Validated GPGMap matches are used to establish loop closure constraints in a pose graph, where submap origins (blue ellipses) are joined by VIO (Visual-Inertial Odometry) pose constraints.}
	\label{fig:overview}
\end{figure}
GPGM-SLAM is built upon a distributed visual SLAM system based on submaps, designed to enable localization and mapping within teams of heterogeneous robots equipped with stereo vision systems \cite{schuster2018distributed}, such as the Lightweight Rover Unit (LRU) rovers \cite{Schuster2017TowardsAP} and the UAV ARDEA \cite{lutz20ardea} developed at the DLR (German Aerospace Center) Institute of Robotics and Mechatronics.
On both systems, stereo images are processed with the Semi-Global Matching (SGM) algorithm \cite{hirschmuller2007stereo} implemented on an FPGA to obtain per-pixel depth information, which is used to evaluate the terrain traversability but also to compute visual odometry.

Local state estimation is performed as a fusion of visual odometry, IMU measurements and, if available, wheel odometry. The fusion is based on a loosely coupled approach and estimates 6D poses as well as sensor biases using an Extended Kalman Filter (EKF), implemented as a Local Reference Filter \cite{schmid2014local}.
It allows long-term operation by enabling a consistent estimation of all the unobservable state variables within the system, i.e. translations in x, y, z and yaw angle, with locally bounded uncertainties. Switching to a new local reference frame is actively triggered by a set of conditions, which include the length of the local trajectory, the accumulated yaw rotation of the robot body, and exceeding a threshold on the state covariances.

Each local reference frame defines also the origin of a submap, which is a local representation of the environment with a bounded size and error due to the aforementioned frame switching conditions.
It represents the geometry of the environment in form of a 3D point cloud as well as an additional probabilistic 3D voxel grid.
Submaps are aggregated from stereo depth measurements using the pose estimated from the local reference filter.
In the original SLAM system from \cite{schuster2018distributed}, submaps distinguish obstacles and traversable parts, and obstacles are not only utilized for local navigation but also to define 3D keypoints to extract C-SHOT features using the normals of the full point cloud. With that, the annotated submap point clouds are utilized for the goal of establishing loop closures by matching 3D features and validating a transformation from Hough3D clusters followed by a 4D ICP step.

Submap matches produce rigid transformation constraints between the associated local reference filters, which define the nodes of a graph optimized by iSAM2 \cite{kaess2012isam2} using the GTSAM library. Subsequent nodes are constrained by inter-pose constraints between the local reference frames defined by the local state estimation. Loop closures constrain non-adjacent nodes, and are followed by an update of the iterative iSAM2 optimizer. The optimized poses are finally used to correct the appearance of the full map, which is built by assembling the rigid submaps' point clouds and voxel grids according to the positions of their local reference frames.

In the course of this paper, we will refer to this configuration of the SLAM system as a baseline for comparison. Furthermore we will also refer to a variation of the system \cite{giubilato2021multimodal}, which refers to an improved formulation of local submaps with the inclusion of visual keyframes captured from the camera images along the local trajectory. In this variation, keyframes are used to compute ORB features on locations of the image with valid depth values. ORB features are finally matched to compute inter-pose constraints between keyframes, which are constrained to the local reference frames from visual-inertial constraints.

The system denoted GPGM-SLAM, see Figure~\ref{fig:overview}, substitutes the point cloud-based submaps with GPGMaps to increase the performances in the loop closure detection stage. GPGMaps are built by leveraging Gaussian Processes to infer a continuous elevation given the submap point cloud as a set of training points. This allows to overcome the traditional issues with the point cloud data structure, e.g. noise and occlusions, to the end of matching on challenging unstructured scenes. Compared to the baseline system, GPGMaps are not only matched given the overlap from the prior on poses, but also using a Bag of Words approach with SIFT features computed on the gradient of the elevation. Each GPGMap also contains the ``parent'' submap to refine the final transformation with a 4D ICP step.
We will present details on construction of GPGMaps in the following section.


\section{Gaussian Process Gradient Maps}
\label{sec:GPGM}

The concept of GPGMap was initially introduced in our previous work \cite{gentil2020gaussian} to address the challenge of noisy and sparse geometric data in the context of submap-based place recognition in unstructured planetary environments.
GPGMaps are a continuous and probabilistic representation of the gradient of the terrain elevation. In this representation a direct prediction of the elevation's derivatives is  inferred from 3D point clouds of the system's environment using GPs and linear operators \cite{Sarkka2011}.
While GP models allow for data-driven accurate interpolation, they suffer from a cubic $\mathcal{O}(n^3)$ computational complexity for the first inference and linear $\mathcal{O}(n)$ for each additional prediction, with $n$ the number of sample data.
In our previous work \cite{gentil2020gaussian}, this constraint was simplistically addressed with a naive index-based downsampling of the submap point clouds.
While improving computation time, this naive method reduces the data information used in the inference.

\begin{figure}
    \centering

    \begin{tikzpicture}
        \node[anchor=south west,inner sep=0] at (0,0) {\includegraphics[clip, width=14cm]{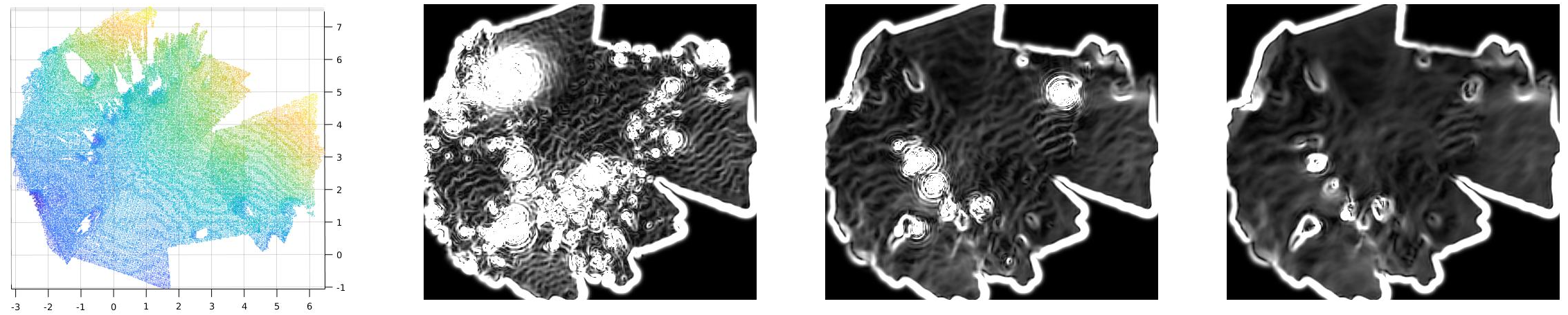}};
	    \node[anchor=south] at (1.6,-0.60) {(a)};
	    \node[anchor=south] at (5.3,-0.60) {(b)};
	    \node[anchor=south] at (8.85,-0.60) {(c)};
	    \node[anchor=south] at (12.4,-0.60) {(d)};
    \end{tikzpicture}
	\caption{Inference of the gradient of the terrain elevation using D-SKI \cite{eriksson2018scaling}. (a) is the input point cloud (colours correspond to different elevation values). (b), (c), and (d) are the D-SKI inferences using both as inputs the elevation information and ad-hoc normals computed using 10, 50, and 100 closest points, respectively.}
    \label{figure:d_ski_fail}
\end{figure}
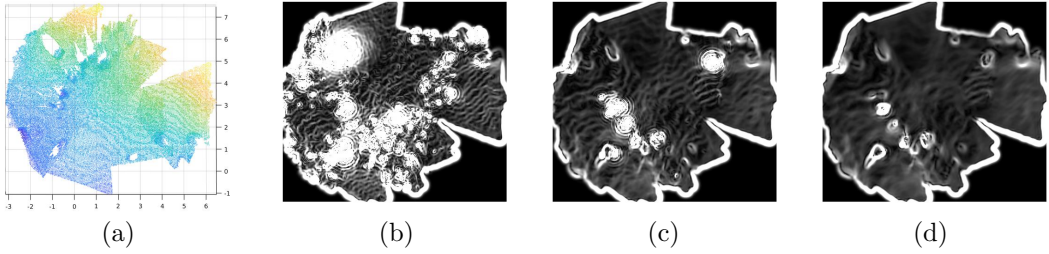

In this work, to improve on the processing time and to allow the use of denser information, we employed a variant with derivatives of the Structured Kernel Interpolation (SKI) scheme \cite{wilson2015kernel}. SKI is a GP-kernel approximation method based on interpolation/inducing points used to alleviate the cubic complexity of regular GP regression. Work on derivatives with SKI-based GP interpolation \cite{eriksson2018scaling}, so-called D-SKI, was later introduced to leverage the SKI scheme and the signal's derivatives (as input) to improve the inference output.
D-SKI allows for the inference of the signal's derivatives but requires observations of the derivatives in the first place.
Unfortunately these derivatives, which in our case are equivalent to the normal vectors of the submap point clouds, are not readily available as an observation.
Instead, if required as input, the normal vectors have to be computed from the point cloud using numerical methods that require manual parameter tuning.
As an example, to generate Figure~\ref{figure:d_ski_fail}, we estimate the submap normal vectors with a closest-neighbor search and a principal component analysis to provide D-SKI GPs with both elevation and derivative information.
By looking at the gradient inferences with different parameters for the closest-neighbor search, it is clear that the output greatly depends on the \emph{ad-hoc} derivative estimation method and the original spatial point distribution.
Consequently, one can conclude that using D-SKI over model-based generated derivatives (as opposed to actual observations of the derivatives) fails the data-driven nature of GP regression and can significantly degrade its performance.
In this work, we propose a variation of the SKI scheme to directly infer the elevation derivatives using elevation measurements solely.
We will further refer to this variation as SKI-D.

\begin{figure}[t!]
\centering
{
\fontfamily{lmss}\selectfont
\includesvg[width=.7\linewidth]{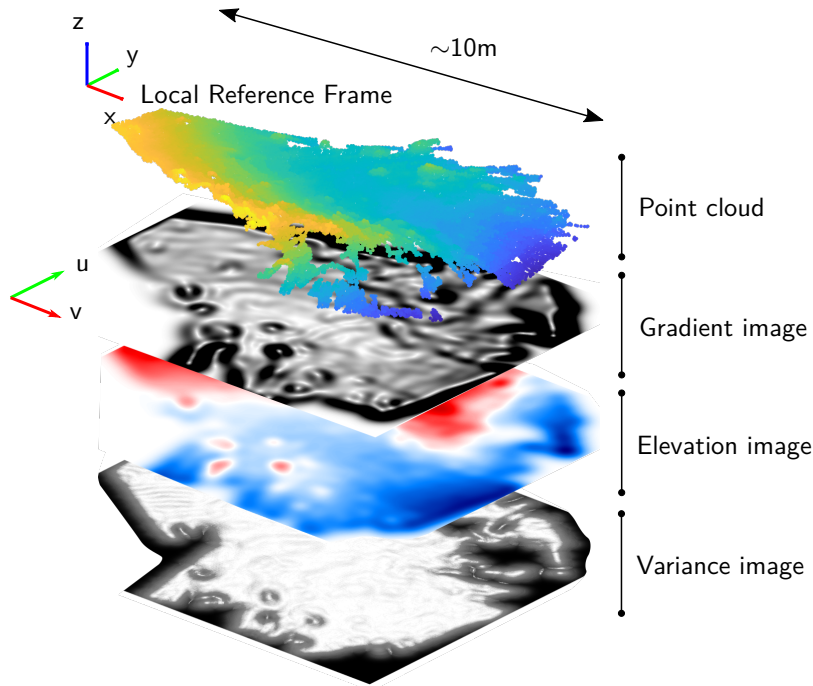}
}
	\caption{Anatomy of a GPGMap: A point cloud, inherited from the parent submap, is used to train a Gaussian Process which infers elevation in a spatially continuous manner using SKI. The colormap of the elevation image highlights negative and positive elevation values, which are normalized on a zero mean. The variance of the elevation is instead approximated given the distance to the closest training point and the length scale of the squared exponential kernel, darker colors suggest higher variances which grow rapidly with the distance from the closest train point (the variance image is trimmed for the purpose of visualization, however the high variance regions extend to infinity in the spatial domain). The gradient image encodes the magnitude of the spatial gradients of the elevation image and is used to compute SIFT features for loop closure detection and matching of GPGMaps.}
\label{fig:gpgmap_anatomy}
\end{figure}

This section first presents the required GP regression and SKI background before detailing the concepts involved in the generation of GPGMaps as shown in Figure~\ref{fig:gpgmap_anatomy}.

\subsection{Gaussian Process regression and SKI}

\newcommand{\kernel}[2]{\mathbf{K}(#1,#2)}

\newcommand{\kernelfunction}[2]{k_f(#1,#2)}

Let us consider a signal $h$, function of $t$, with a GP \cite{Rasmussen2006} as
\begin{align}
	& h(t) \sim \mathcal{GP}\big(0,k_h(t,t')).
\end{align}
The function $k_h(t,t')$ is called the kernel function and represents the covariance between two instances of the signal $h$:
\begin{align}
	& \text{cov}\Big(h(t), h(t')\Big) = k_h(t,t').
\end{align}

Given noisy $N$ observations of that signal
\begin{align}
	& y_i = h(t_i) + \eta_i, \ \ \ \ \eta_i \sim \mathcal{N}\big(0,\sigma_y^2), \ \ \ \ (i = 1,\cdots, N),
\end{align}
the mean and variance of $h$ at any input $t$, $h^*(t)$ and $\sigma_h^*(t)$ respectively, can be inferred as
\begin{align}
	h^*(t) &= \mathbf{k}_h(t,\mathbf{t}) \Big[\mathbf{K}_h(\mathbf{t},\mathbf{t}) + \sigma_y^2 \mathbf{I} \Big]^{-1} \mathbf{y},
	\label{eq:gp_regression}
	\\
	\sigma_h^*(t) &= k_h(t,t) - \mathbf{k}_h(t,\mathbf{t}) \Big[\mathbf{K}_h(\mathbf{t},\mathbf{t}) + \sigma_y^2 \mathbf{I} \Big]^{-1}\mathbf{k}_h(t,\mathbf{t})^\top,
\end{align}
with $\mathbf{y}$ the column vector of the noisy observations $\begin{bmatrix}y_1,\cdots,y_N\end{bmatrix}^\top$, $\mathbf{k}_h(t,\mathbf{t})$ the row vector of covariances of $h$ between the inference input and the observations' inputs $\begin{bmatrix}k_h(t,t_1),\cdots,k_h(t,t_N)\end{bmatrix}$, and $\mathbf{K}_h(\mathbf{t},\mathbf{t})$ the covariance matrix of the observations $\begin{bmatrix}\mathbf{k}_h(t_1,\mathbf{t})^\top,\cdots,\mathbf{k}_h(t_N,\mathbf{t})^\top\end{bmatrix}$.

As discussed above, the main drawback of GP regression is its computational complexity of $\mathcal{O}(n^3)$ due to the matrix inversion in \eqref{eq:gp_regression}.
The SKI principle \cite{wilson2015kernel} relies on the use of inducing points $U$ arranged as a regular grid.
By formulating the approximation $\mathbf{K}_{\mathbf{t},U} \approx \mathbf{W}_\mathbf{t}\mathbf{K}_{U,U}$ with $\mathbf{W}_\mathbf{t}$ a matrix of interpolation weights, it is possible to approximate $\mathbf{K}_{\mathbf{t},\mathbf{t}} = \mathbf{K}_h(\mathbf{t},\mathbf{t}) + \sigma_y^2 \mathbf{I}$ as
\begin{align}
	\mathbf{K}_{\mathbf{t},\mathbf{t}} \approx \mathbf{W}_\mathbf{t}\mathbf{K}_{U,U}\mathbf{W}_\mathbf{t}^\top = \mathbf{K}_{\text{SKI}}.
\end{align}
By choosing a local interpolation method, the matrix $\mathbf{W}_\mathbf{t}$ can be extremely sparse and the inversion $\mathbf{K}_{\text{SKI}}^{-1}\mathbf{y}$ is solved using a linear conjugate gradient only requiring matrix vector products.
Similarly the covariance vector $\mathbf{k}_h(t,\mathbf{t})$ is computed with a fixed amount of operations as $k_h(t,t') \approx \sum\limits_{i} w_i(t)k_h(u_i, t')$.
It results in a computationally efficient method to infer the signal's mean for any input $t$: linear time $\mathcal{O}(n)$ inference with $\mathcal{O}(1)$ for additional predictions.

\subsection{SKI-D for Gaussian Process Gradient Maps generation}
Unless in the context of specific exploration tasks (e.g. exploration of caves or lava tubes \cite{leveille2010lava}), the environment surrounding the rover is unlikely to contain overhanging or vertical structures.
Accordingly, we assume that the environment can be modeled as a 2.5D map without losing significant structural details.
Given that submaps are gravity-aligned, their elevation, or the $z$ component of the point cloud, can be expressed as a function $\elevationfunction{\inputdomain{}}$ with $\inputdomain{}$ the $x$ and $y$ spatial coordinates within the submap's local reference frame.
Considering the elevation function modelled with a zero-mean GP as
\def\abscissa{\mathbf{x}}
\begin{align}
    &\elevationfunction{\inputdomain{}} \sim \mathcal{GP}\big(0,\kernelfunction{\inputdomaini{}{\i}}{\inputdomaini{}{\i'}} \big),
    \nonumber
    \\
    \z{}{\i} &= \elevationfunction{\inputdomaini{}{\i}} + \eta_{\i}, \ \ \ \ \eta_{\i} \sim \mathcal{N}(0,\sigma^2_{z}),
    \label{eq:gp_model}
\end{align}
and leveraging the application of linear operators for GP regression \cite{Sarkka2011}, the derivatives of the elevation can exactly be inferred solely as a function of the noisy elevation observations:
\begin{align}
	\frac{\partial \elevationfunction{\inputdomain{}}}{\partial x} = \frac{\partial \mathbf{k}_f(\inputdomain{},{\inputdomainmat})}{\partial x} \big[ \mathbf{K}_f(\inputdomainmat,\inputdomainmat) + \sigma^2_{z} \mathbf{I}\big]^{-1}\mathbf{\z},
    \nonumber
    \\
	\frac{\partial \elevationfunction{\inputdomain{}}}{\partial y} = \frac{\partial \mathbf{k}_f(\inputdomain{},{\inputdomainmat})}{\partial y} \big[ \mathbf{K}_f(\inputdomainmat,\inputdomainmat) + \sigma^2_{z} \mathbf{I}\big]^{-1}\mathbf{\z}.
\end{align}

As per \cite{eriksson2018scaling}, based on the SKI formulation, the differentiation operator can be applied to the kernel as $\frac{\partial k_f(\mathbf{x},\mathbf{X})}{\partial x} \approx \sum\limits_{i} \frac{\partial w_i(\mathbf{x})}{\partial x}k_f(\mathbf{u}_i, \mathbf{X})$ and $\frac{\partial k_f(\mathbf{x},\mathbf{X})}{\partial y} \approx \sum\limits_{i} \frac{\partial w_i(\mathbf{x})}{\partial y}k_f(\mathbf{u}_i, \mathbf{X})$.
Consequently, the elevation's derivatives are approximated as
\begin{align}
	\frac{\partial \elevationfunction{\inputdomain{}}}{\partial x} \approx \Big(\frac{\partial \mathbf{w}(\mathbf{x})}{\partial x} \mathbf{K}_{U,U}\Big) \mathbf{K}_{\text{SKI}}^{-1}\mathbf{\z},
	\label{eq:gp_derivative_x}
    \\
	\frac{\partial \elevationfunction{\inputdomain{}}}{\partial y} \approx \Big(\frac{\partial \mathbf{w}(\mathbf{x})}{\partial y} \mathbf{K}_{U,U}\Big) \mathbf{K}_{\text{SKI}}^{-1}\mathbf{\z}.
	\label{eq:gp_derivative_y}
\end{align}

The gradient component of the GPGMaps is then obtained as the norm of the derivatives:
\begin{align}
	g(\mathbf{x})~=~\sqrt{\Big(\frac{\partial \elevationfunction{\inputdomain{}}}{\partial x}\Big)^2 + \Big(\frac{\partial \elevationfunction{\inputdomain{}}}{\partial y}\Big)^2}.
	\label{eq:gp_gradient}
\end{align}

\subsection{Elevation's variance approximation}

Unfortunately, the SKI scheme as presented in \cite{wilson2015kernel} and \cite{eriksson2018scaling} does not allow for efficient inference of the signal's variance.
In order to provide a proxy for the variance of the elevation inferences, we leverage the kernel function between an inference input $\mathbf{x}$ and the centroid of the noisy observations that are present in an arbitrary radius around the inference input.
For efficiency, a k-d tree is used for the neighbor search in the $x$-$y$ plane.
If no point is present in the radius search, the closest neighbor is used instead of the neighbors' centroid.
Formally, the approximation of the variance is defined as $\tilde{var}(f(\mathbf{x})) = k_f(\mathbf{x}, \tilde{\mathbf{x}})$, with $\tilde{\mathbf{x}}$ being the centroid of the neighbors in a certain radius around $\mathbf{x}$ in the noisy observation's input $\mathbf{X}$, or simply the closest observation input if none is present in the radius search.

\section{Loop Closure with GPGMaps}
\label{sec:LC_detection}
Establishing loop closures in our SLAM system is accomplished by attempting pairwise matching of GPGMaps.
The procedure leverages the representation of elevation, variance, and gradient of elevation as images to use elements of standard image registration pipelines.
More specifically, as new GPGMaps are created, SIFT features are extracted on the gradient image.
To avoid extracting features on uncertain regions, where the elevation is inferred far from neighboring training points, the variance image is used to establish a binary mask.
Let the i-th GPGMap of a mapping session be denoted as:
\begin{equation}
	\mathcal{G}_i = \langle \mathbf{T}_\text{LRF}^w, P, I, V, G, K, F \rangle_i
\end{equation}
where $\mathbf{T}_\text{LRF}^w$ is the pose of Local Reference Frame for the i-th GPGMap with respect to a world coordinate frame. $P$ is the original point cloud, $I$ and $V$ are elevation and variance images, $G$ is the elevation gradient, $K$ and $F$ represent the sets of keypoints and feature descriptors.
The task of establishing matches between GPGMaps is split in two main parts. First, a preliminary candidate selection is made relying either on appearance or prior pose information.
Second, the candidate matches are validated by attempting to fit an SE(2) transformation between gradient images.
The relative pose between the local reference frames of matched GPGMaps enforces loop closure constrains in the SLAM graph.

\subsection{Loop Closure Detection}
As new GPGMaps are created, the ensemble of SIFT descriptors computed on the gradient image is translated into a lightweight representation using the Bag of Words paradigm.
More specifically, we use a modified version of the DBoW2 \cite{GalvezTRO12} library introduced in \cite{giubilato2020relocalization}.
By means of a vocabulary of visual words, built appropriately considering the potential features that can appear during execution, the Bag of Words model in DBoW2 summarizes feature occurrences into a map of indexes and corresponding weights, called BoW vector.
The weighting scheme is based on the \textit{tf-idf} (i.e., term frequency–inverse document frequency) metric, which weights the effectiveness of a word to suggest a visual context depending on the number of times the word was observed in the images used to build the vocabulary.
That ensures that common features have little impact on the similarity metric between BoW vectors.
Finally, an additional weight is applied to the feature occurrences to account for the incompleteness of the vocabulary.
Vocabularies built on sparse and partial data might not fit well the operational scenario, and the feature descriptors parsed to the vocabulary might lay far from the closest word in the descriptor space.
This distance is accounted for, as explained in more detail in \cite{giubilato2020relocalization}, to downweight the contribution of the word to the similarity of BoW vectors.

Let be $\mathbf{v}_i$ the BoW vector for GPGMap $\mathcal{G}_i$:
\begin{eqnarray}
    \mathbf{v}_i & = & \{\langle \text{id}_0, \bar{w}_0\rangle, ..., \langle \text{id}_j, \bar{w}_j\rangle  \} \\ \nonumber
    \bar{w}_j & = & w^*_j \cdot idf \cdot tf \nonumber
\end{eqnarray}
where $\text{id}_j$ is the index of the j-th word and $\bar{w}_j$ is the corresponding weight, which is computed as the product of all weighting factors including $w^*_j$ which encodes the distance to the closest visual word and $idf \cdot tf$ which is the tf-idf term.
All new GPGMaps are compared with a database of the GPGMaps previously built from the beginning of the session to identify potential matches.
The similarity of GPGMaps is computed in a fast and approximate manner by comparing the BoW vectors through a similarity score $s(\mathbf {v}_i, \mathbf {v}_{db})$, computed as in \cite{GalvezTRO12}:
\begin{equation}
	s(\mathbf {v}_i, \mathbf {v}_{db}) = 1 - \frac{1}{2}\bigg| \frac{\mathbf{v}_i}{|\mathbf{v}_{i}|} - \frac{\mathbf{v}_{db}}{|\mathbf{v}_{db}|} \bigg| \quad \in [0, 1]
\end{equation}
For every GPGMap, we select the two most similar ones and attempt match validation.

In addition, all GPGMaps are associated with a Local Reference Frame, the pose of which is added to the SLAM graph.
Thus, it is possible at all times to use the prior pose information to select matching candidates based on spatial overlap.
This is not only the strategy employed by our baseline SLAM system \cite{schuster2018distributed} but also from the majority of structure-based SLAM systems \cite{ebadi2020lamp,mendes2016icp} usually relying on LiDAR point clouds.
Specifically, we compute a bounding box for each GPGMap considering the extend of the original point cloud.
The bounding box is then inflated considering the variance on the pose coordinates.
For every new GPGMap, we compute the Intersection over Union (IoU) of its bounding box with the GPGMaps in the database.
Furthermore, we monitor the status of the pose graph, maintained by the SLAM node, and in case significant changes are observed the IoU is recomputed for all pairs of GPGMaps.
Non-matched and sufficiently overlapping GPGMaps are then selected as candidates for match validation.

A match queue is built by adding first the candidate matches based on BoW similarity and secondly the candidates from overlap.
This way, in case a match of the first type is validated, the pose graph is immediately updated correcting poses and covariances of all GPGMaps.
This allows to more accurately compute the overlap metric and, as covariances decrease, to reduce the number of candidates to be validated, reducing the computational overhead.

\subsection{Loop Closure Validation}
In this step of the pipeline, candidate GPGMap matches are validated to select true matches and compute a final rigid transformation between the original point clouds that is used to constrain their local reference frames in the SLAM graph.
The validation is articulated in two steps. A first stage concerns the gradient images and, more specifically, their SE(2) alignment through SIFT features matching. A second stage consists in deriving the final transformation between the original point clouds given a prior alignment from the first step.

\subsubsection{SE(2) Alignment of Elevation Images}
\begin{figure}[ht!]
\centering
\includesvg[width=.5\linewidth]{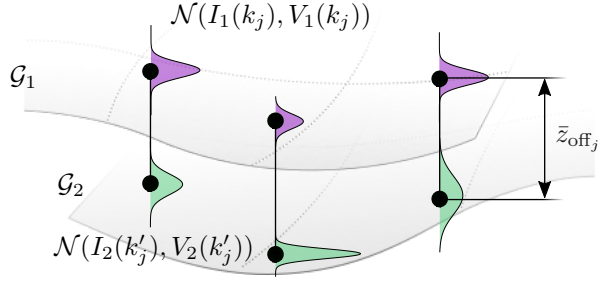}
\caption{Estimation of the $z$ offset from the aligned elevation $I$ and variance $V$ images of GPGMaps $\mathcal{G}_1$ and $\mathcal{G}_2$. The SE(2) transformation for the alignment is obtained from SIFT correspondences $k_j$ and $k_j'$ computed on the gradient images.}
\label{fig::z_off}
\end{figure}
The SIFT features $f_1$ and $f_2$ of a candidate GPGMap pair $(\mathcal{G}_1, \mathcal{G}_2)$ are matched to determine an SE(2) transformation between the gradient images $G_1$ and $G_2$.
Feature matching is performed either in a brute force manner or using the ratio test \cite{lowe2004distinctive}.
The matching strategy is selected by the user at runtime.
In our experiments we find that the ratio test allows to efficiently reject outliers.
However, in the more challenging sequences, the stricter check results in losing good matches compared to the simpler brute force approach, performed in both directions.
The choice between the two can be therefore made empirically depending on the appearance of the environment.
The resulting set of matching features is used to fit an SE(2) transformation between the images in a RANSAC scheme as explained in \cite{gentil2020gaussian}.
A first check is performed on the number of inliers, which must be higher than a predefined threshold. This value is preliminarily set to 5 and is generally valid for all test datasets.

Compared to \cite{gentil2020gaussian}, to reduce the computational overhead, we here accept the result of RANSAC based solely on the elevation of matched features instead of the whole elevation images.
Let be $I_1(k_j)$ and $I_2(k_j')$ be elevations corresponding to the locations of the matched SIFT keypoints in GPGMaps $\mathcal{G}_1$ and $\mathcal{G}_2$.
Let also $V_1(k_j)$ and $V_2(k_j')$ be the associated variances.
Having computed the SE(2) transformation that aligns $I_1$ to $I_2$, the goal here is to determine the remaining parameter to align the point clouds in the 3D space, which is the $z$ offset. This offset is due to the fact that elevations are determined relatively to the local reference frame of $\mathcal{G}_1$ and $\mathcal{G}_2$, leading to potential differences between the local elevations of matching SIFT keypoints.
Furthermore, as a variance is known over the elevation, every offset can be represented as a Gaussian distribution:
\begin{equation}
z_{\text{off}, j} = \mathcal{N}(I_1(k_j) - I_2(k_j'), V_1(k_j) + V_2(k_j'))
\end{equation}
For all $n$ correspondences, the offset that best aligns in $z$ the two GPGMaps can be estimated as:
\begin{equation}
\bar{z}_{\text{off}} = \frac{
\sum_{j=1}^{n} \frac{I_1(k_j) - I_2(k_j')}{V_1(k_j) + V_2(k_j')}
} {
\sum_{j=1}^{n} \frac{1}{V_1(k_j) + V_2(k_j')}
}
\end{equation}
which is the average of all offsets weighted on their variance, or the results of weighted least squares regression for a constant function.

After applying $\bar{z}_{\text{off}}$ to the elevation, we perform a final check to ensure that the pair of GPGMaps is a true match. This is done by examining the Bhattachayyya distances between the aligned elevations $z_1$ and $z_2$:
\begin{eqnarray}
    z_1 &=& \mathcal{N}(I_1(k_j)-\bar{z}_{\text{off}}, V_1(k_j)) \\ \nonumber
    z_2 &=& \mathcal{N}(I_2(k_j), V_2(k_j)) \\ \nonumber
    D_B(z_1,z_2) &=& \frac{1}{4} \ln \left ( \frac 1 4 \left( \frac{\sigma_1^2}{\sigma_2^2}+\frac{\sigma_2^2}{\sigma_1^2}+2\right ) \right ) +\frac{1}{4} \left ( \frac{(\mu_1-\mu_2)^{2}}{\sigma_1^2+\sigma_2^2}\right )
\end{eqnarray}
If the GPGMap match is true, the $z$ offset should determine low Bhattacharyya distances for all keypoint pairs. Therefore we accept the match if more than 70\% of the pairs lead to $D_B(z_1,z_2)<t_{D_B}$ where $t_{D_B}$ is set to 2 from empirical considerations.

\subsection{Point Cloud Alignment}
Having validated a match between a pair of GPGMaps, a transformation between the original point clouds has to be derived in order to establish a loop closure constraint in the pose graph. A transformation $\mathbf{T}_{\mathcal{G}_2}^{\mathcal{G}_1} \in SE(2)$ is known from the last step as well as an offset $\bar{z}_{\text{off}}$ that aligns the elevations $I_1$ to $I_2$. A transformation in the 3D space from the Local Reference Frames of the original point clouds is obtained as:
\begin{eqnarray}
\mathbf{T}_{LRF_1}^{LRF_2} &=& \mathbf{T}_{\mathcal{G}_2}^{\text{LRF}_2} \ \mathbf{T}_{\mathcal{G}_1}^{\mathcal{G}_2} \ \mathbf{T}_{\text{LRF}_1}^{\mathcal{G}_1} \\
&=&
\begin{bmatrix}
\mathbf{I} &
\begin{matrix}
    rx_o \\
    ry_o \\
    0 \\
\end{matrix} \\
\begin{matrix}
0 & 0 & 0
\end{matrix} & 1
\end{bmatrix}^{-1}
\
\begin{bmatrix}
\mathbf{R}(\phi) &
\begin{matrix}
    rt_x \\
    rt_y \\
    \bar{z}_{\text{off}} \\
\end{matrix} \\
\begin{matrix}
0 & 0 & 0
\end{matrix} & 1
\end{bmatrix}^{-1}
\
\begin{bmatrix}
\mathbf{I} &
\begin{matrix}
    rx_o' \\
    ry_o' \\
    0 \\
\end{matrix} \\
\begin{matrix}
0 & 0 & 0
\end{matrix} & 1
\end{bmatrix}
\end{eqnarray}
where $\mathbf{R}(\phi)$ is the rotation component of $\mathbf{T}_{\mathcal{G}_1}^{\mathcal{G}_2}$ with roll and pitch angles set to zero, $r$ is the resolution of elevation and variance images (m/pix), and $x_o$ and $y_o$ are the origins of the elevation and variance images in the Local Reference Frame of the original submaps.
To account for potential misalignments after validation, we perform an 4D ICP on the point clouds using $\mathbf{T}_{LRF_1}^{LRF_2}$ as the initial transformation. Finally, the RMSE error of the point-to-point distances at the last iteration of ICP is used to approximate a covariance for the refined transformation. This transformation is then used to establish a loop closure constraint in the pose graph.

\section{Evaluation of Loop Closure Detection}
\label{sec:bow_eval}
\begin{figure}[t!]
\centering
{
\fontfamily{lmss}\selectfont
\includesvg[width=\linewidth]{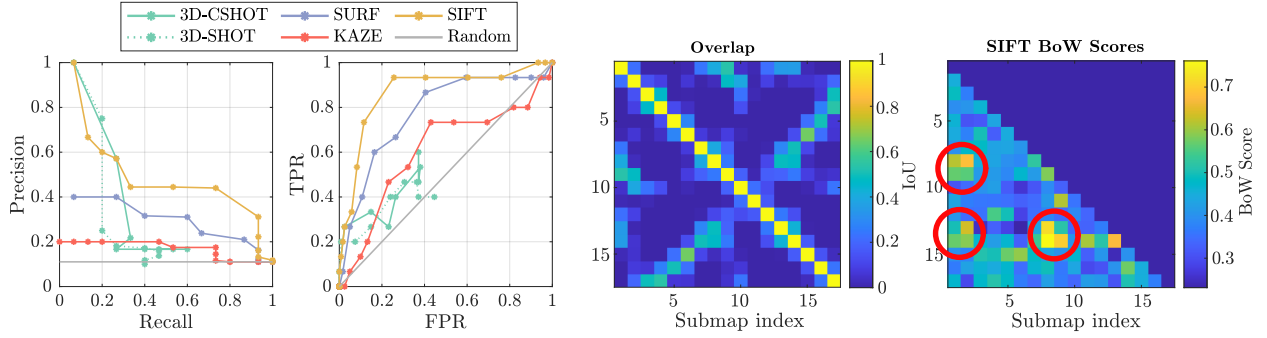}
}
\caption{Comparison of the classification performances of matching GPGMap pairs using different visual feature descriptors in a BoW scheme with traditional 3D feature matching in a RANSAC scheme. The results refer to the etna\_03 sequence. In the top row, a precision-recall plot as well as a ROC (Receiver Operating Characteristic Curve) illustrate the different performances of the evaluated classifiers compared to the performances of a random selection. The bottom row shows heatmaps of the spatial overlap between GPGMaps compared with the pairwise scores of BoW vectors built using SIFT descriptors. Red circles highlight examples of true loop closures corresponding to high BoW scores and overlap.}
\label{fig:bow_eval}
\end{figure}

We perform here a first evaluation of the performances that can be expected in recollecting similar GPGMaps from a database.
As visual place recognition using the Bag of Words paradigm is very mature and has been implemented in many visual SLAM systems, we aim to evaluate the feasibility of utilizing the same techniques and apply them to the structure-oriented case.
Furthermore, this preliminar assessment also aims at motivating the choice of the visual descriptor which we utilize in GPGM-SLAM given the superior classification performances.

Here we implement a Bag of Word framework in the MATLAB environment, in order to easily analyze various descriptor choices.
Although the performances might differ from the DBoW2 library that we use in GPGM-SLAM, we expect them to be comparable and, most of all, we expect the different descriptors to exhibit a similar behaviour between the two implementations.
We evaluate the SIFT \cite{lowe1004distinctive}, SURF \cite{bay2006surf} and KAZE \cite{alcantarilla2012kaze} feature descriptors, extracted on the gradient images from the Etna sequences, see the dataset presentation in Section~\ref{sec:eval}.
A vocabulary of features is constructed using the gradient images from a subset of GPGMaps built while evaluating GPGM-SLAM on the sequences.
To every GPGMap, a single BoW vector is associated given the set of descriptors extracted on the relative gradient image.
Higher scores between BoW vectors denote higher similarity between the GPGMaps, and we compare it with the overlap of the bounding boxes of the original point clouds given the prior pose from SLAM.
This, in fact, can be generally considered accurate enough to discriminate potentially matching GPGMaps between all the possible pairs.

Figure~\ref{fig:bow_eval} reports precision-recall and ROC (Receiver Operating Characteristic Curve) curves built by varying a threshold between the minimum and maximum BoW scores to classify GPGMap pairs as matching.
True GPGMap matches are instead selected using the Intersection over Union (IoU) metric between bounding boxes, choosing empirically a threshold to discriminate between matching and non-matching.
Among all the evaluated visual descriptors, SIFT is the one which obtains the highest classification performances. Note that these performances are related to non-validated candidate GPGMap matches. We expect, in fact, that the precision after validation will be 100\% in order to avoid validating any false GPGMap match. Overall, the SIFT descritptor allows here to select a wider set of true GPGMap matches and therefore facilitating the task of detecting numerous loop closures along the trajectory.
Figure~\ref{fig:bow_eval} also compares the overlap of GPGMaps, or IoU values, and the BoW scores between all GPGMap pairs, highlighting with red circles three examples of correct loop closure detections.

We also compare the performances of a traditional approach for structure registration, using the original point clouds that are used to generate the GPGMap. 3D-SHOT and 3D-CSHOT refer to matching keypoints from high curvature regions \cite{giubilato2020relocalization} using the SHOT and CSHOT descriptors \cite{salti2014shot} followed by RANSAC to determine a relative 6D rigid transformation given pairs of matching 3D keypoints.
The precision-recall curves for these configurations are determined by varying the distance threshold between descriptors to get keypoint matches between the two point clouds. As the precision-recall curves show, the structure-based approach is able to correctly select a small set of matching pairs and the performances degrade with an immediate loss of precision for low recall values. Note that this alternative is order of magnitudes more computationally expensive than evaluating the similarity between BoW vectors and therefore completely infeasible for larger GPGMap sets. Furthermore, the approach depends on the possibility of extracting 3D keypoints which, as will be later discussed, is the main limitation of structure-based approaches in our target scenarios.

\section{Performance Evaluation of SLAM System in Planetary-like Environments}
\label{sec:eval}
\begin{table}[ht!]
\centering
\footnotesize
\caption{Presentation of the Etna sequences}
\label{table:sequences}
\begin{tabularx}{\linewidth}{|lllcX|}
\hline
\Tstrut Sequence         & Time [min] & Length [m] & Example Image & Description  \\
\hline\hline
\Tstrut etna\_01         &  8            & 30           & \raisebox{-\totalheight}{\includegraphics[width=1.2in]{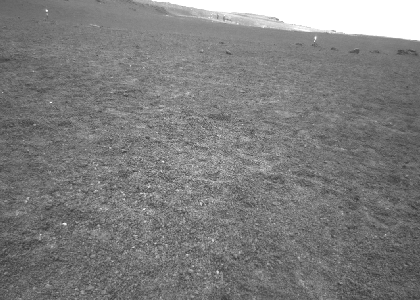}} &
		- Short rectilinear travel forward and back to the start. Rover is teleoperated
		\newline- Sandy, featureless flat ground \Tstrut
        \\
 etna\_02         & 65            & 820           & \raisebox{-\totalheight}{\includegraphics[width=1.2in]{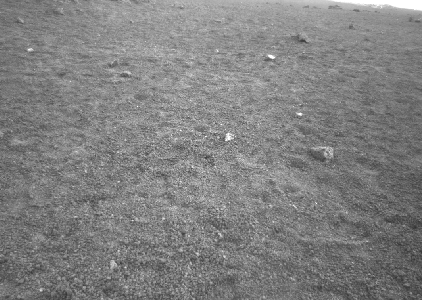}} &
 - Long distance travel. Rover is teleoperated, pan-tilt mechanism not used, camera pointing to the same direction \newline- Sandy and hilly terrain, does not contain rocks or obstacles \\
 etna\_03         &  25            & 125           & \raisebox{-\totalheight}{\includegraphics[width=1.2in]{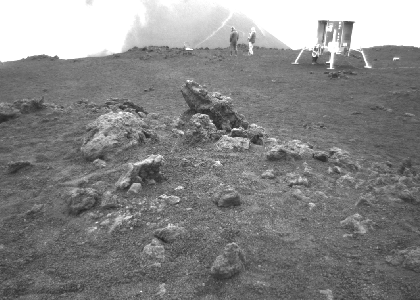}} & - Rover performs waypoint-based autonomous navigation, using often the pan-tilt mechanism to assess traversability. \newline- Hilly terrain containing plenty of visual and structural features   \\
 etna\_04         &  39            & 112           &  \raisebox{-\totalheight}{\includegraphics[width=1.2in]{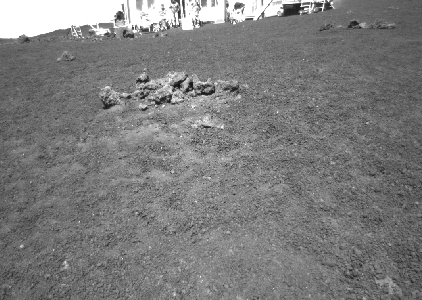}} & - Autonomous exploration sequence, no user intervention. The rover moves frequently using the pan-tilt mechanism, therefore jerky camera motions pose challenges to the state estimation \newline- Traversable terrain, few rocks scattered and observed at a safe distance  \\
etna\_05         &  40            & 170          & \raisebox{-\totalheight}{\includegraphics[width=1.2in]{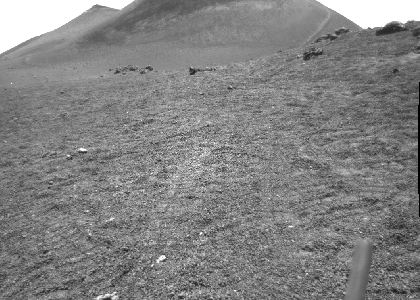}} & - As etna\_04                   \\
 \hline
\end{tabularx}
\end{table}

\begin{table}[ht!]
\centering
\footnotesize
\caption{Presentation of the Morocco sequences}
\label{table:sequences_morocco}
\begin{tabularx}{\linewidth}{|lllcX|}
\hline
\Tstrut Sequence         & Time [min] & Length [m] & Example Image & Description  \\
\hline\hline
\Tstrut E0  &      
      15    &     223       & \raisebox{-\totalheight}{\includegraphics[width=1.2in]{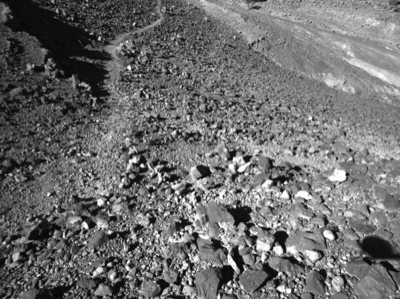}} & - Traverse on a hiking path through a rocky slope. SUPER frequently stops and looks around. \newline- Traversing same exact path on the way back from the opposite direction, start and end of the trajectory coincide. \newline- Rough terrain with big stones.   \\
	 E1  &   
      16           &    309        & \raisebox{-\totalheight}{\includegraphics[width=1.2in]{fig/impressions/kess_1/1.png}} & - As E0 \\
	 E2  &     
      25      &    374        & \raisebox{-\totalheight}{\includegraphics[width=1.2in]{fig/impressions/kess_1/1.png}} & - Same environment as E0 and E1. Big traverse off the hiking path with little to no overlap. Return on the initial spot while traversing on previously observed places. \\ \hline
\end{tabularx}
\end{table}

We evaluate the localization performances of GPGM-SLAM on two datasets recorded in planetary-like environments on Mt. Etna, Sicily and in the Morocco desert. The first dataset \cite{vayugundla2018datasets} was recorded in 2017 during the final demonstration mission of the Helmholtz project ROBEX, where the volcanic environment of Mt. Etna appears as a lunar landscape. The lack of visual and structural features, along with the repetitiveness of the visual appearance, makes place recognition a hard task, challenging the ability of robots to recognize previously visited places and correct localization drift. Similarly, the second dataset, named MADMAX and collected on the Moroccan desert \cite{lukas2021MADMAX} aims at evaluating localization algorithms in challenging outdoor environments. The appearance of this dataset is similar to the one of Martian landscapes, and covers a wide variety of scenarios. A summary of the sequences used to test GPGM-SLAM is given in Table~\ref{table:sequences} and Table~\ref{table:sequences_morocco}.

GPGM-SLAM is evaluated in a comparative performance analysis with our baseline SLAM system, in two configurations, as well as the plain output of the visual-inertial odometry. A global trajectory is assembled by transforming in world coordinates the local trajectories that belong to each submap, or GPGMap. Let be $\mathbf{p}^{\mathcal{G}_i} = \{p_j, \ j=1,..,n \}^{\mathcal{G}_i}$ trajectory of the camera belonging to GPGMap i. The full trajectory at the end of a SLAM session can be reconstructed through the optimized poses from the pose graph $\mathbf{T}_{\mathcal{G}_\text{i}^\text{w}}$. The reconstructed trajectory is aligned to a D-GPS ground truth leveraging temporal correspondences. The first corresponding positions between trajectory and ground truth are fixed, and Horn's quaternion-based method is used to find the rotation matrix that ensures the best fit. We then evaluate the Root Mean Square (RMS) error between corresponding positions to define a final performance metric:
\begin{equation}
    RMSE = \sqrt{\frac{1}{n_c}\sum_{j=0}^{n_c}||p_j - p_j^\text{D-GPS}||^2}
\end{equation}
Furthermore, we compute the final error, or the absolute trajectory error evaluated at the final correspondence with the D-GPS track, as well as the relative errors which are normalized with respect to the trajectory length.
For all sequences and for each compared configuration of the SLAM system we perform 5 runs to account for non-deterministic effects due to RANSAC and scheduling of processes in the operative system. The algorithms are tested on a laptop equipped with an Intel i7-4810MQ CPU and 16 Gb of RAM, whose computational performances are on par with the PC mounted on the LRU rover.

\subsection{Mt. Etna Datasets}
Sequences from the Mt. Etna datasets are recorded from the perspective of the LRU rover while performing a variety of tasks involving different levels of autonomy and environments. More specifically, the etna\_02 sequence is recorded while imparting direct commands to the rover through a joystick with the aim of traversing a long path with two overlapping loops to evaluate long-term localization \cite{grixa2018appearance}. As the rover is fully operated, the pan-tilt mechanism of the camera head is not used, and cameras are generally pointed towards the ground. This, together with the locally straight trajectory, result in very narrow submaps which contain little structure (see impression in Table~\ref{table:sequences}) and are very challenging to match. A similar type of environment is observed in etna\_01 although in this sequence, together with etna\_03, the rover is navigating autonomously across waypoints. In this case, the obstacle avoidance mechanism requires to frequently observe the environment in the proximity of the rover, which causes submaps to both have a wider extent and include more obstacles and peculiar structures. The etna\_03 sequence is the easiest sequence of the set for the purpose of establishing loop closures through submap and GPGMap matching. The rover follows here a trajectory shaped as a ``figure 8'' with two long overlapping sequences of about 15 meters length, the first on a rocky area and on the same driving direction, the second on a flat and sandy area and traversed from the opposite direction. In the etna\_04 and etna\_05 sequences, the rovers move in a fully autonomous manner as a tradeoff between exploring a scene, covering as much ground as possible, and revisiting places \cite{lehner2017exploration}. These two sequences are the most challenging, in this evaluation, for the purpose of loop closure detection. Fully autonomous behaviors require the rovers to acquire as much information as possible about the environment, this results in frequent and jerky motions of the camera head. This limits the spatial extent of submaps, as new ones are more frequently triggered by rapid increase in the covariance of local pose estimation. In order to maximize traversability, the environment also contain very little amount of rocks and patterns on the sand to recognize. Furthermore, unlike the other datasets, the trajectories from etna\_04 and etna\_05 do not overlap after long distances (e.g. start to end) but contains only few and short term loops, which limits the effectiveness of establishing loop closures to the overlall localization accuracy. However, the challenging nature of the appearance of the environment allows to benchmark the ability of detecting hard to find loop closures.

\begin{table}[t!]
\centering
\footnotesize
\caption{Results for the Etna dataset averaged on 5 runs. Relative errors are normalized on the approximate length of the trajectory. Matches are average number of validated submap matches}
\label{table:etna_results}
\begin{tabular}{|c|cccc|cccc|cccc|}
\hline
Dataset      & \multicolumn{12}{c|}{ETNA}                                                                                                                                                        \\
\hline
Sequence     & \multicolumn{4}{c|}{etna\_01}                                         & \multicolumn{4}{c|}{etna\_02}                                          & \multicolumn{4}{c|}{etna\_03}  \\
\hline
Config       & filter & 3D   & 3D-KF & GPGM                                            & filter & 3D    & 3D-KF & GPGM                                          & filter & 3D   & 3D-KF & GPGM                                                            \\
\hline
RMSE (avg)   & 0.44   & 0.50 & 0.48  & \textbf{0.41}                                  & 9.39   & 7.88  & 9.06  & \textbf{2.84}                                  & 0.92   & 0.44 & 0.37  & \textbf{0.34}                                                            \\
\hline
EndErr (avg) & 0.98   & 1.23 & 1.18  & \textbf{0.67}                                  & 32.26  & 26.44 & 30.37 & \textbf{10.95}                                 & 1.05   & 1.34 & 1.00  & \textbf{0.88}                                                            \\
\hline
RMSE (\%)    & 0.75   & 0.87 & 0.82  & \textbf{0.71}                                  & 1.14   & 0.96  & 1.10  & \textbf{0.35}                                  & 0.75   & 0.36 & 0.30  & \textbf{0.27}                                                            \\
\hline
MaxErr (\%)  & 1.68   & 2.11 & 2.03  & \textbf{1.15}                                  & 3.93   & 3.22  & 3.70  & \textbf{1.33}                                  & 0.85   & 1.08 & 0.81  & \textbf{0.70}                                                            \\
\hline
Matches      & -   & - & -  & \textbf{0.83}                                  & -   & 0.25  & -  & \textbf{19.4}                                 & -   & 1.2 & 3  & \textbf{8}                                                            \\
\hline
\end{tabular}

\vspace{0.5cm}

\begin{tabular}{|c|cccc|cccc|}
\hline
Dataset      & \multicolumn{8}{c|}{ETNA}                                                                                \\
\hline
Sequence     & \multicolumn{4}{c|}{etna\_04}                                          & \multicolumn{4}{c|}{etna\_05}  \\
\hline
Config       & filter & 3D   & 3D-KF & GPGM                                            & filter & 3D   & 3D-KF & GPGM                                                            \\
\hline
RMSE (avg)   & 1.50   & 1.48 & 1.30  & \textbf{1.25}                                   & 3.01   & 3.04 & 3.03  & \textbf{1.97}                                                            \\
\hline
EndErr (avg) & 2.94   & 2.80 & 2.55  & \textbf{2.27}                                   & 7.73   & 7.75 & 7.85  & \textbf{4.98}                                                            \\
\hline
RMSE (\%)    & 1.33   & 1.31 & 1.15  & \textbf{1.11}                                   & 1.78   & 1.79 & 1.79  & \textbf{1.16}                                                            \\
\hline
MaxErr (\%)  & 2.60   & 2.48 & 2.25  & \textbf{2.01}                                   & 4.56   & 4.57 & 4.63  & \textbf{2.93}                                                            \\
\hline
Matches      & -   & - & 2.2  & \textbf{4.6}                                   & -   & - & -  & \textbf{2}                                                            \\
\hline
\end{tabular}
\end{table}
\begin{figure}[t!]
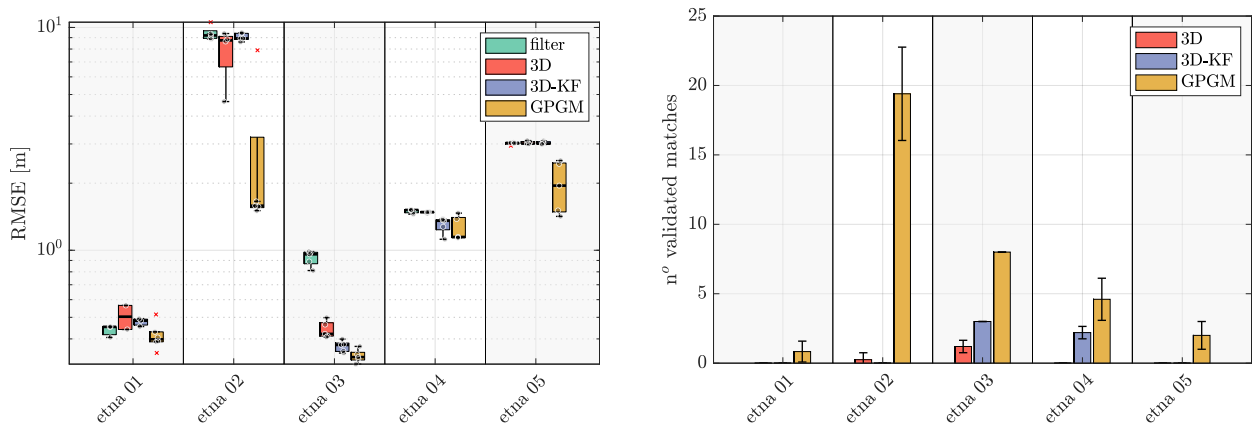

	\centering
	\begin{subfigure}[b]{.475\textwidth}
        \includesvg[width=\linewidth]{fig/new_numbering/rmse_etna_all}
	\end{subfigure}\hfill
	\begin{subfigure}[b]{.475\textwidth}
        \includesvg[width=\linewidth]{fig/new_numbering/match_etna_all}
	\end{subfigure}
	\caption{RMSE errors and submap or GPGMap matches for the Etna sequences}
	\label{fig::rmse}
\end{figure}

Results for the Etna sequences are summarized in Table~\ref{table:etna_results}, where GPGM-SLAM is compared with our baseline SLAM system as well as the visual-inertial odometry alone (i.e. no submap matching), denoted ``filter". Bold faces denote the best results for all the sequences, and it is immediately apparent how GPGM-SLAM outperforms all configurations, delivering better fitting trajectories to the D-GPS track and therefore lower maximum and final errors.

As all SLAM configuration share the same inputs, i.e. submaps, the sole reason for the higher performances of GPGM-SLAM is the ability to establish more, and more sparsely distributed, constraints in the pose graph from GPGMap matches.
Results are particularly impressive for the etna\_02 and etna\_01 cases, where submaps do not contain rocks or evident structures in general.
In fact, on etna\_01 no submaps are matched from CSHOT correspondences and on etna\_02 only one match happens and only in 1 run out of 5.
GPGM-SLAM instead matches consistently around 20 GPGMaps on etna\_02, when the rover drives on the same path for the second time, as well as 1 to 2 GPGMaps on etna\_02 out of a total of 8 submaps in the sequence.
Figure~\ref{fig:match_tiles} displays a variety of GPGMap match examples including the alignment of gradient images and point clouds.
Specifically for etna\_01, the lack of structure is immediately apparent also from the point cloud view, where rocks emerging from the sand are shaded in darker colors.
Submap pairs from etna\_02 have a similar appearance to the ones from etna\_01 although being narrower.
The etna\_03 sequence is the easiest of the set, giving many opportunities to match either submaps or GPGMaps.
The trend of RMSE errors from Figure~\ref{fig::rmse}, in fact, show decreasing errors for the 3D, 3D-KF and GPGM configurations.
The environment from etna\_03 is rocky and the rover frequently repeats the same path also observing the scene from similar perspectives.
This facilitates place recognition from multiple modalities, i.e. visual and structural perception, favoring the 3D-KF case over the 3D only configuration.
Both of the approaches, however, are only able to match submaps from the parts of the sequence richer of structure, where also the rover revisits the place from the same driving direction. GPGM-SLAM instead, matches submaps also in the parts of the sequence where the rover revisits from the opposite driving direction, due to the inherent benefit of Gaussian Process regression to infer elevation where it is missing (e.g. obstructed from direct line of sight).
Finally, for etna\_04 and etna\_05, although obtaining lower errors than the baseline, performances of GPGM-SLAM in terms of absolute localization are not impressive due to the pose drift accumulated by jerky camera motions which are not compensated by the lack of long range loop closures. On etna\_05, the most challenging sequence on the dataset, the baseline SLAM system is unable to perform any submap matches, while GPGM-SLAM is able to match a few GPGMaps and partially compensate for localization drift.
\begin{figure}[htp]
	\centering
	{
	\fontfamily{lmss}\selectfont
	\includesvg[width=\linewidth]{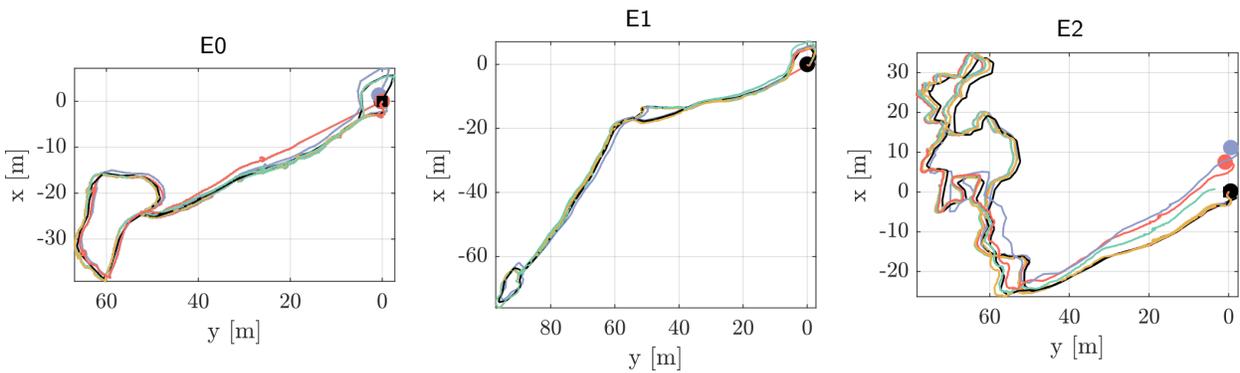}
	}
	\caption{Trajectories for all sequences and algorithms on the Etna and Morocco datasets. Squares and circles highlight respectively the start and end of the trajectories}
	\label{fig::trajs_etna}
\end{figure}
In Figure~\ref{fig::trajs_etna} all the best fitting trajectories are plotted against the D-GPS ground truth, reflecting the different behaviour of the rover as previously stated.

To contextualize the performance figures with respect to the state of the art of Visual SLAM systems, we attempted to evaluate ORB-SLAM2 \cite{Mur-Artal2017} and ORB-SLAM3 \cite{campos2020orb}.
However, as cameras are mostly looking downward towards the sandy ground, and the visual appearance is highly ambiguous and aliased, both algorithms frequently lose tracking of ORB features.
For ORB-SLAM2 this translate into failure, unless the camera revisits similar viewpoints. For ORB-SLAM3 instead, this translates into frequent creation of new maps, which the place recognition scheme is unable to match.
As fragments of the reconstructed trajectories are too short to be evaluated (less than a couple of meters in general), we refer/postpone their evaluation to the Morocco dataset, where the camera motion from the hand-held unit facilitates visual tracking.

\begin{figure}[htp]
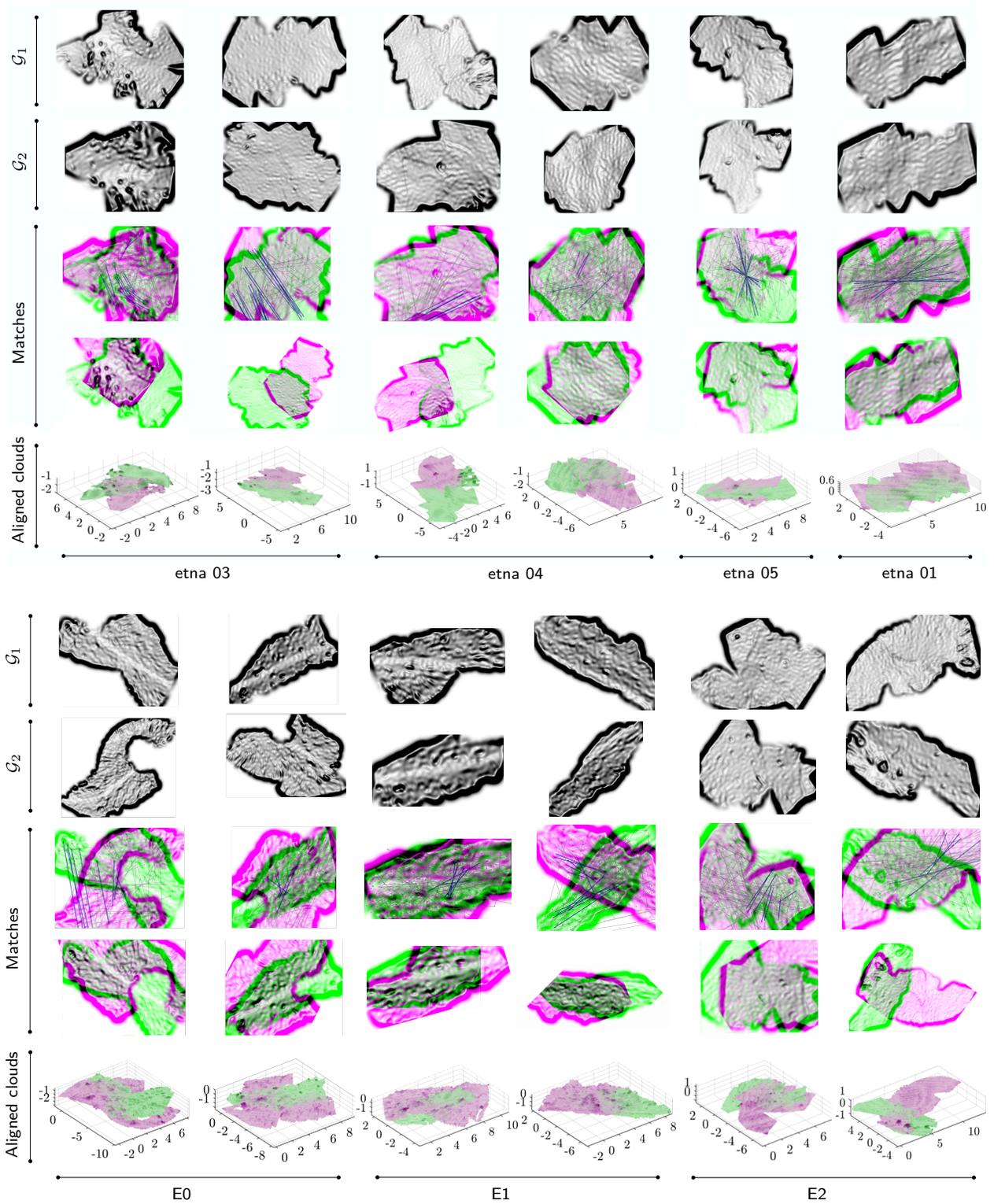

    \centering
    \begin{subfigure}[b]{\textwidth}
    {
    \fontfamily{lmss}\selectfont
    \footnotesize
    \includesvg[width=\linewidth]{fig/new_numbering/match_tiles_downsampled2}
    }
    \end{subfigure} \\ \vfill
    \begin{subfigure}[b]{\textwidth}
    {
    \fontfamily{lmss}\selectfont
    \footnotesize
    \includesvg[width=\linewidth]{fig/new_numbering/match_tiles_morocco}
    }
    \end{subfigure}
    \caption{Impressions of GPGMap matches for the Etna (top) and Morocco (bottom) datasets. $\mathcal{G}_1$ and $\mathcal{G}_2$ denote the gradient images of the GPGMap pair, "Matches" denotes validated SIFT feature matches as well as aligned gradients. The last row shows the original point clouds from submaps aligned in the 3D space.}
    \label{fig:match_tiles}
\end{figure}

\subsection{Morocco Datasets}
\begin{table}[t]
\centering
\footnotesize
\caption{Results for the Morocco dataset averaged on 5 runs, single runs for ORB-SLAM2 and VINS. Relative errors are normalized on the approximate length of the trajectory.}
\begin{tabular}{|c|cccc|cccc|cccc|}
\hline
Dataset & \multicolumn{12}{|c|}{MOROCCO} \\
\hline
Sequence & \multicolumn{4}{|c|}{E0} & \multicolumn{4}{|c|}{E1} & \multicolumn{4}{|c|}{E2} \\
\hline
Config &       filter & GPGM & ORB & VINS &              filter & GPGM & ORB & VINS &            filter & GPGM & ORB & VINS \\
\hline
RMSE (avg) &   0.61 & \textbf{0.50} & 1.52 & 0.82 &      1.52 & \textbf{0.88} & - & 1.00 &       2.34 & \textbf{1.82} & 3.02 & 3.83 \\
\hline
EndErr (avg) & 1.14 & 0.33 & \textbf{0.19} & 1.29 &      3.42 & 0.48 & - & \textbf{0.03} &       6.45 & \textbf{2.93} & 7.44 & 11.36 \\
\hline
RMSE (\%) &    0.25 & \textbf{0.21} & 0.68 & 0.37 &      0.46 & \textbf{0.27} & - & 0.32 &       0.54 & \textbf{0.42} & 0.75 & 1.01 \\
\hline
MaxErr (\%) &  0.46 & 0.14 & \textbf{0.09} & 0.57 &      1.03 & 0.15 & - & \textbf{0.01} &       1.49 & \textbf{0.68} & 1.85 & 2.99 \\
\hline
Matches &      -    & 6.60 & - & - & - & 7.40 & - & - & - & 2.50 & - & - \\
\hline
\end{tabular}
\label{table:morocco_results}
\end{table}
\begin{figure}[ht!]
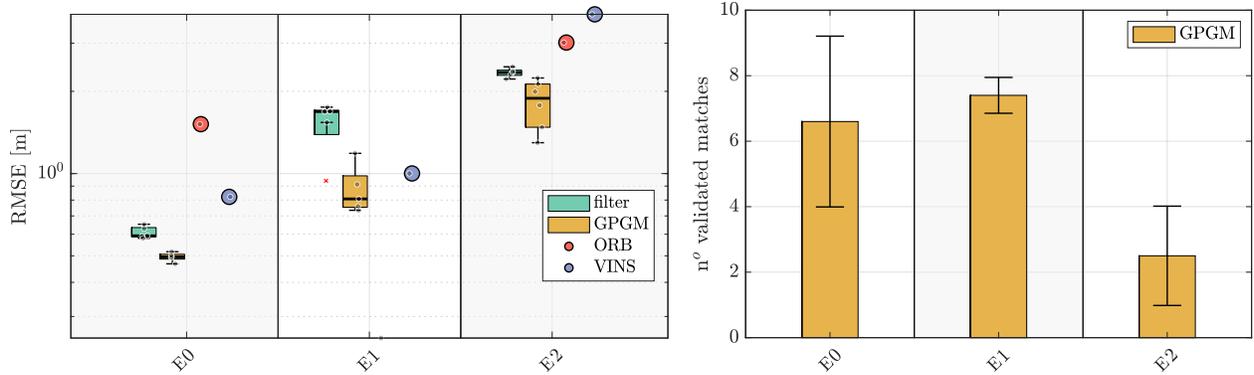

	\centering
  \begin{subfigure}[b]{.53\textwidth}
        \includesvg[width=\linewidth]{fig/new_numbering/rmse_morocco_all}
	\end{subfigure}\hfill
	\begin{subfigure}[b]{.45\textwidth}
        \includesvg[width=\linewidth]{fig/new_numbering/match_morocco_all}
	\end{subfigure}
	\caption{RMSE errors and submap or GPGMap matches for the Morocco sequences}
	\label{fig:rmse_morocco}
\end{figure}
The sequences denominated E0, E1 and E2 take place on and off an hiking trail located on a rocky slope in the ``Kess Kess'' formations, Erfoud area, Morocco.
The sequences are named, for consistency, as within the released MADMAX dataset \cite{lukas2021MADMAX}.
Compared to the sequences from the Etna dataset, the environment on these sequences is scattered with rocks, see the impressions from Table~\ref{table:sequences_morocco}, whose uniformity again challenges the ability to recognize specific places.
Furthermore, compared to the frequent panning motions of the LRU rover camera, in all the sequences from the Morocco dataset, the camera usually points forward along the direction of motion resulting in narrower GPGMaps and therefore comprising less structural details.
As an additional challenge, while returning to the starting location, the path is traversed from the opposite direction.
This compromises the ability of performing place recognition using visual cues and we specifically evaluate this in Section~\ref{sec:gpgmap_vs_ibow}.

We compare the performances of GPGM-SLAM here against our visual-inertial localization as a baseline, as well as with ORB-SLAM2 and VINS-MONO in order to contextualize our results with well known algorithms from the SLAM community.
For reasons of consistency, we utilize the trajectories estimated from ORB-SLAM2 and VINS as released as part of the MADMAX dataset.
A single trajectory from both algorithms is provided for every sequence, resulting from evaluating both with an optimal parameter set.
Table~\ref{table:morocco_results} reports the errors of all tested algorithms and configurations, which are graphically represented in Figure~\ref{fig:rmse_morocco}.
For the Morocco dataset we avoided to report the performances of our baseline configuration 3D and 3D\_KF, already evaluated on the Etna sequences.
The opposite viewpoints combined with the uniformity of the structure, lead to a significant degree of occlusions and complementarity between potentially matchable submaps.
Therefore due to the lack of shared surface details and unique 3D keypoint detections, submap matching on the level of point clouds show poor performances and often producing establishing wrong loop closures. We consider therefore these approaches unable to produce satisfactory results on the Morocco sequences and we avoid to report performance metrics for them.

In all the Morocco sequences GPGM-SLAM produces the closest trajectory to the D-GPS ground truth, achieving RMSE errors as low as the 0.2\% of the trajectory length.
This is due to the many GPGMap matches established along the overlapping parts of the trajectories, even if traversed from the opposite direction. Examples of matched GPGMaps are collected in Figure~\ref{fig:match_tiles}.
Interestingly, although both ORB-SLAM2 and VINS obtain higher RMSE errors, in E0 and E1 they score the lowest final errors due to the fact that they detect loop closures at the very end of the sequences, where the camera observes objects and equipment from similar perspectives.
The inability of detecting visual loop closures within the trajectory, however, does not permit to recover the inevitable pose drift which degrades the fit to the D-GPS track.
Results of ORB-SLAM2 are not reported for E1 as it was unable to reconstruct more than approximately 25\% of the trajectory due to non recovered tracking loss.

\subsection{SIFT features design choices}

Using SIFT features on GPGMaps corresponds to the detection of corner-like keypoints on the elevation's gradient associated with descriptors computed based on the gradients of the elevation's gradient.
This set-up aims to justify empirically the choice of using SIFT over elevation's gradient compared to other variations.

The standard procedure to match two images using SIFT features consists in computing differences of Gaussians at different scales on both images to extract salient keypoints.
Then, descriptors based on orientation histograms of the intensity gradients are computed over scaled patches around each keypoints.
Technically, it requires the algorithm to calculate the gradient of the images numerically.
Finally, the descriptors are used to match features between the images using various distance functions and matching strategies.
Regarding this pipeline and the availability of both the elevation and elevation's gradients as image-like data (using GP regression without or with differentiation linear operators), our first design choice concerns the data used for keypoint detection; should the keypoints be extract over the terrain elevation or its gradient?

Let us consider five true loop closures in the dataset etna\_03 with given SE(2) transformations between each GPGMap pairs.
Keypoints are extracted in each submap using independently both the elevation and the elevation's gradient.
For each submap pair, the keypoints of one submap are projected in the second submap using the associated SE(2) transformation.
Then, keypoint pairs across maps are classified as true matches or not according to a distance threshold of three centimeters.
Using the elevation for keypoints extraction results in an average of 5.8 true matches per loop closure, while the gradient-based keypoints results in an average of 11.2 true matches.
These results coincide with the intuition that the elevation's gradient presents more corner-like patterns than the elevation data itself.
The rest of this experiments will only consider keypoints detected over the gradient image of the GPGMaps.

Given the availability of the GP-inferred derivatives of the elevation, \eqref{eq:gp_derivative_x} and \eqref{eq:gp_derivative_y}, one can consider using these analytical derivatives instead of the numerical differentiation of the elevation's gradient performed when using standard SIFT features over GPGMaps.
For each keypoint, we computed the SIFT descriptors using independently both the GP-inferred elevation derivatives (i) and the numerical gradient of the elevation's gradient as per the classic use of SIFT features (ii).
Note that these approaches are fundamentally different as (i) uses $\frac{\partial \elevationfunction{\inputdomain{}}}{\partial x}$ and $\frac{\partial \elevationfunction{\inputdomain{}}}{\partial y}$, while (ii) uses $\frac{\partial g(\inputdomain{})}{\partial x}$ and $\frac{\partial g(\inputdomain{})}{\partial y}$ with $g(\inputdomain{})$ defined in \eqref{eq:gp_gradient}.\footnote{The derivatives of $g(\inputdomain{})$ cannot be directly inferred using linear operators over GP kernels as per the non-linear operations in \eqref{eq:gp_gradient}}
We analysed the matching distances for inliers and outliers using both techniques.
The results are shown in Table~\ref{table:sift_experiment}.
The significantly bigger difference between inlier and outlier distances for the standard SIFT descriptors confirms that the gradient of the elevation's gradient (ii) offers more distinctive features and lead to better loop-detection performances.
Additionally, the extra computation needed for the numerical differentiation is not penalising the real-time abilities of our system as shown in the following section (the descriptors computation represents less than 3\% of the total GPGMap generation and processing time).

\begin{table}[ht]
\centering
\footnotesize
	\caption{L1 distances between feature descriptors using the GP-based gradient or the standard SIFT over the elevation's gradient (leveraging numerical differentiation over the elevation's gradient).}
\begin{tabular}{|c||c|c|c|}
\hline
	\scriptsize Method & \scriptsize Avg inlier desciptor dist. & \scriptsize Avg outlier descriptor dist. & \scriptsize Nb loop detected
	\\
	\hline
	\hline
	\scriptsize (i) SIFT variant using GP derivatives & 1656.7 & 1704.0 & 2
\\
\hline
	\scriptsize (ii) Standard SIFT over elevation's gradient & 1692.1 & 2337.3 & \textbf{5}
	\\
\hline
\end{tabular}
\label{table:sift_experiment}
\end{table}

\subsection{Timings}
\begin{table}[ht]
\centering
\footnotesize
\caption{Times (in seconds) for every step of GPGM-SLAM averaged on all datasets.}
\begin{tabular}{|c||c|}
\hline
Step & Time (avg) [s] \\
\hline\hline
create gpgpmap & 6.22 \\
\hline
pose callback & 0.01 \\
\hline
compute descriptors & 0.19 \\
\hline
compute bow vector & 0.01 \\
\hline
compute scores & 0.01 \\
\hline
(avg. / Submap) & \textbf{6.43} \\
\hline\hline
match validation: RANSAC & 0.01 \\
\hline
match validation: ICP & 0.62 \\
\hline
\end{tabular}
\label{table:timings}
\end{table}
Table~\ref{table:timings} reports the execution time for all the principal parts of GPGM-SLAM involved during the processing of GPGMaps. All times are referred to single thread implementations, except during GP inference of elevation where 2 threads are used, and are measured while evaluating the pipeline using a laptop. Nevertheless, the times are comparable, if not equal, to those that can be expected from the computers onboard the LRU rovers.
The most computationally expensive step is the creation of GPGMaps. The highest contributions are in order the Gaussian Process inference from SKI (with 1 centimeter resolution) and the computation of the approximate variance.
The second highest time is spent on the extraction of SIFT features while all the remaining tasks, including generation of BoW vectors and retrieval from the database, are performed in less than 10 milliseconds.
It must be noted that the extent of submaps greatly affects the inference time for the elevation: frequent panning motions of the camera head lead to accumulating points in the original submap in all directions, therefore growing the set of training points for the Gaussian Process.
For larger submaps, the time of generating GPGMaps can reach up to 5 or 6 seconds in the tested sequences.
The computation time must, however, be compared with the time span of an individual submap which generally varies from 20 to 60 seconds while the camera is moving.
Therefore, the generation of GPGMap from submaps only takes a small fraction of the temporal extent of submaps, inducing a small and relatively constant delay between concluding the creation of one submap and performing loop closure detection with the respective GPGMap.
In terms of match validation, which only occurres while evaluating candidate matches, fitting an SE(2) transformation between gradient images requires a negligible amount of time.
The highest contribution to the validation time is the 4D ICP step on the original point clouds, which takes in average about 0.6 seconds.

\subsection{Comparison with visual relocalization}\label{sec:gpgmap_vs_ibow}
\begin{figure}[htp]
\centering
{
\fontfamily{lmss}\selectfont\footnotesize
\includesvg[width=\linewidth]{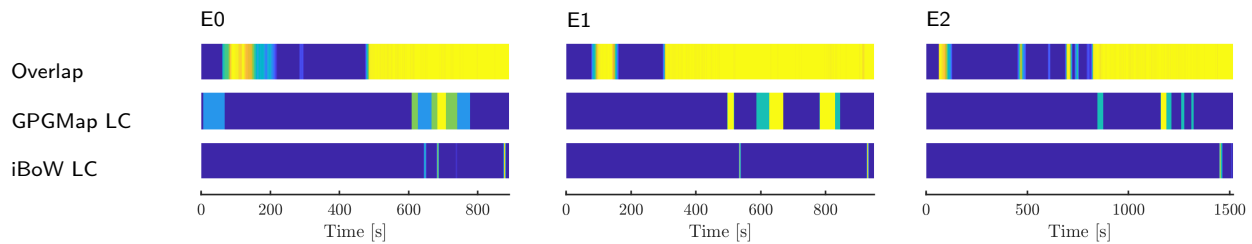}
}
\caption{Frequency of loop closure occurrences for GPGM-SLAM and the visual loop closure detector iBoW-LCD. Heatmaps show the frequency over 5 runs for each algorithm, to account for randomness. The first field, denoted overlap, is proportional to the proximity of the camera for every time step to a past location, if closer than 5 meters. For GPGM-SLAM, a single loop closure occurrence covers the time interval of the relative GPGMap, for iBoW-LCD a single loop closure occurrence is related to a keyframe match and slightly inflated for visualization purposes.}
\label{fig:occ_matches}
\end{figure}

We compare the loop closure detection performances of GPGM-SLAM with the ones of a state of the art visual loop closure detector, iBoW-LCD \cite{garcia2018ibow}. This method is a recent incremental loop closure detector which eliminates the need of a vocabulary of visual word and exploits the concept of visual islands to group consequent images and reduce the computational time to detect a loop closure. The authors report higher retrieval performances than DBoW2, used by ORB-SLAM2, and therefore we use it to evaluate the different behavior of visual-only methods to detect loop closures on our challenging datasets.

We implement iBoW-LCD extracting a maximum of 1500 ORB features per image, using the left frame of the stereo setup. Loop closure validation is performed by fitting a fundamental matrix using RANSAC as implemented in the OpenCV library. We consider a visual loop closure successful if a minimum of 30 inliers is selected after RANSAC. All parameters are selected empirically such that the chance of establishing false loop closures is minimized.

\begin{figure}[t!]
\centering
\includesvg[width=\linewidth]{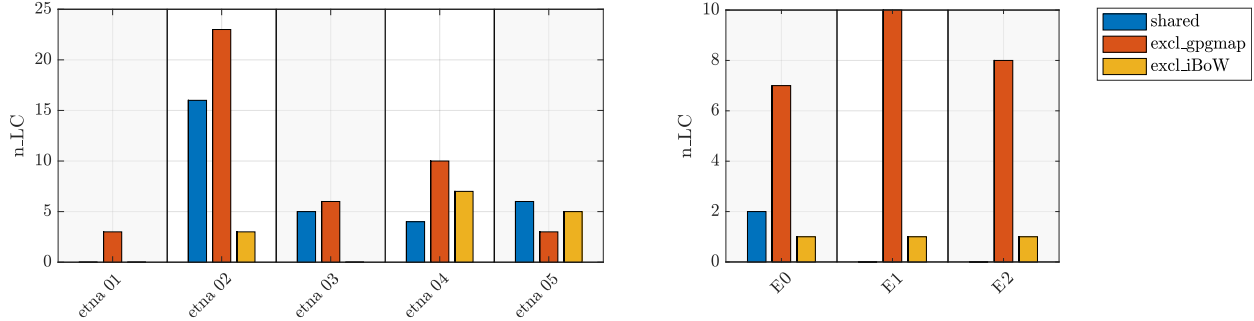}
\caption{Number of shared or exclusive occurrences of loop closures between GPGM-SLAM and iBoW-LCD. One occurrence of visual loop closure from iBoW-LCD is determined by the presence of any number of frame matches within the time interval of a GPGMap. Occurrences are either shared, if both algorithm procude a loop closure for the same GPGMap interval, otherwise exclusive for one of them.}
\label{fig:n_lc}
\end{figure}

In Figure~\ref{fig:occ_matches} we report the frequency of loop closures from both GPGM-SLAM and iBoW-LCD. For GPGM-SLAM, loop closures consists in validated GPGMap matches and their occurrence is registered in the time frame of the query GPGMap, or the one that is matched with one from the past. For iBoW-LCD an occurrence of visual loop closure is registered as a sparse event in a specific time. The time refers to the extent of the D-GPS track. We visually compare the frequencies of loop closure detections with an overlap value. The overlap is computed proportionally to the proximity of the current camera pose, taken from the D-GPS ground truth, with the closest one from the past trajectory.
Figure~\ref{fig:n_lc} reports the number of shared and exclusive loop closures from GPGM-SLAM and iBoW-LCD. As loop closures are defined differently from the two algorithms, we consider an occurrence for iBoW-LCD the presence of any frame matches within the time span of an individual GPGMap. Occurrences of loop closures are either ``shared'' if they happen within the time frame of the same GPGMap, otherwise they are ``exclusive'' for one of the two methods. A higher number of exclusive matches denotes a higher loop closure detection power.

The aim of this evaluation is to highlight the different properties of both sensing modalities, i.e. structural properties for GPGM-SLAM and visual appearance for iBoW-LCD. Specifically, we want to evaluate the benefits of matching structure instead of images in unstructured scenarios where the observer path is unconstrained, but also highlight the limitations of performing loop closure detection with a single modality.

\begin{figure}[t!]
	\centering
	\begin{subfigure}[b]{0.5\textwidth}
		\includegraphics[width=\linewidth]{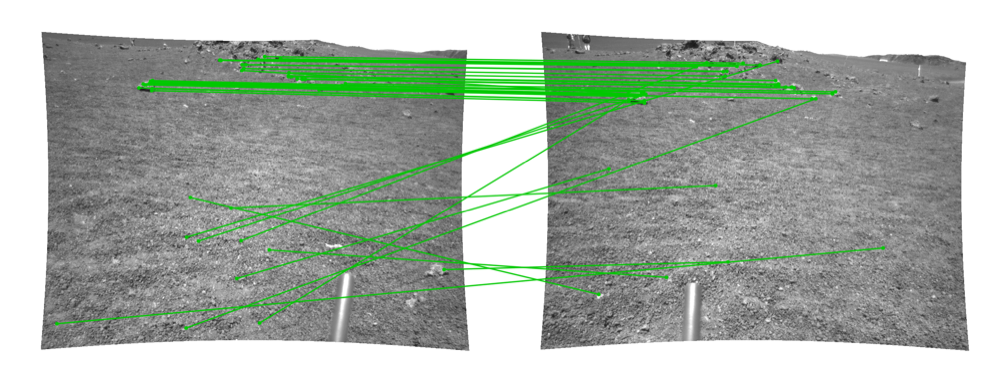}
	\end{subfigure} \\
	\begin{subfigure}[b]{0.5\textwidth}
		\includegraphics[width=\linewidth]{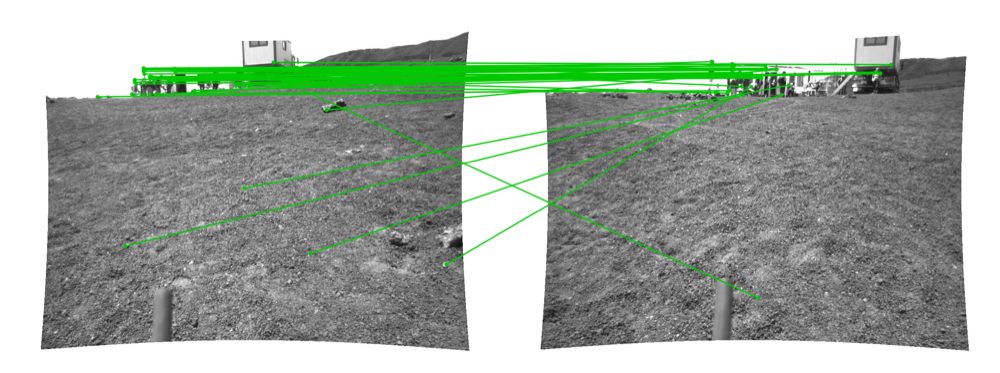}
	\end{subfigure} \\
	\begin{subfigure}[b]{0.5\textwidth}
		\includegraphics[width=\linewidth]{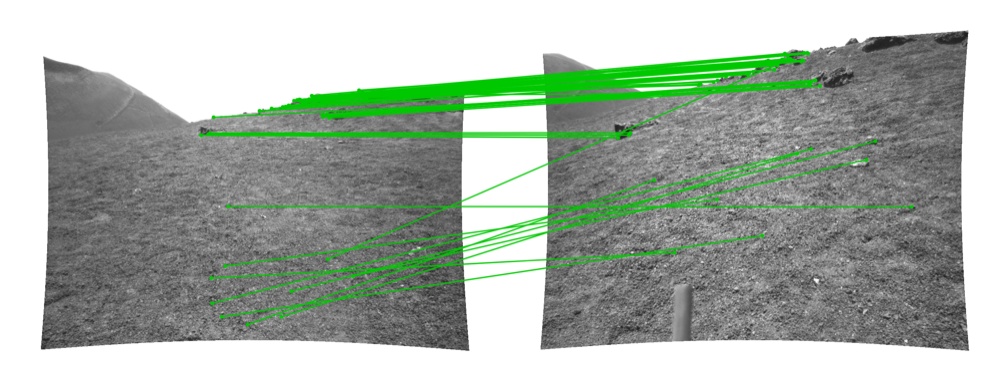}
	\end{subfigure}
	\caption{Examples of loop closures detected by iBoW-LCD within the time span of GPGMaps that were not matched by GPGM-SLAM. Visual features driving the detection of a loop closure lie far from the stereo sensing range, which limits the spatial extent of GPGMaps. Note how wrong feature matches happened in the proximity of the cameras, on sandy terrains. These examples suggest the benefit of embedding multiple modalities to detect loop closures in the context of SLAM.}
	\label{fig:match_ibow_exclusive}
\end{figure}

Although the environment on the Etna sequences is mostly sandy with a only a few clusters of rocks scattered around, the frequent panning of the camera leads to frequently observing the surrounding landscape at a distance. On etna\_01, the rover mostly observes a featureless ground without using the pan-tilt mechanism, and returns to the starting position from the opposite direction. In this case only GPGM-SLAM is able to match GPGMaps and close loops, while iBoW-LCD does not detect any. Differently, on etna\_03 and etna\_02, iBoW-LCD detects many loop closures as the rover revisits frequently the same places from very close positions and viewpoints. However, GPGM-SLAM is able to match more numerous portions of the trajectory and because of that, matches are either shared between the two algorithms or detected exclusively by GPGM-SLAM, see Figure~\ref{fig:n_lc}. The different modality of iBoW-LCD favors it in th etna\_04 and etna\_05 sequences, where many visual loop closure happen in places where GPGM-SLAM does not detect any similarity between GPGMaps, see the high number of exclusive loop closures for iBoW-LCD from Figure~\ref{fig:n_lc}. iBoW-LCD, in fact, detect loop closures given image features which lie far beyond the range of stereo measurements, a few examples are given in Figure~\ref{fig:match_ibow_exclusive}.

The sequences from the Morocco datasets are challenging for visual loop closure detection, as most of the overlapping parts of the trajectory are traversed from the opposite direction. While the elevation, inferred from the Gaussian Process, is continuous and its gradient offers repeatable features regardless of the camera viewpoint, the appearance of the images is completely different. This explains the higher performances of GPGM-SLAM in this case, or the high number of exclusive matches reported in Figure~\ref{fig:n_lc}.

The presence of exclusive loop closure detections from iBoW-LCD in both the Etna and Morocco datasets suggests that multiple modalities could be exploited to improve the chances of relocalization. GPGMap matches infact rely on the proximity of the environment to the camera, as the input point clouds are built from stereo depth, while visual loop closures can happen while observing landmarks far from the sensing range of a stereo setup.

\section{Conclusions}
\label{sec:conclusions}
In this paper we presented GPGM-SLAM, a submap-based SLAM system for mobile robots based on Gaussian Process Gradient Maps targeted at stereo vision systems.
The continuous elevation inferred by a Gaussian Process is used to compute gradient images which contain sufficient and repeatable visual information to make structure-oriented place recognition possible in challenging self similar and ambiguous environments.
Real time operations are permitted by Structured Kernel Interpolation (SKI) which reduce the computational load for inferring the local elevation by orders of magnitude compared to the standard GP formulation with $O(n^3)$ time complexity.
We tested GPGM-SLAM on sequences from two datasets captured on Mt. Etna, Sicily and in the Morocco desert, both planetary-like environments, demonstrating higher performances than our baseline SLAM system \cite{schuster2018distributed} as well as state of the art visual SLAM systems.
Through a specific analysis of the loop closure detection capabilities of our approach compared to visual approaches, we demonstrated that the performance improvements are principally due to the ability of matching GPGMaps produce while traversing on opposite directions.
We in fact find out that the slow and unconstrained motion of mobile robots while operating in the field, often constrained by task specific requirements such as obstacle avoidance and traversability estimation, prevents the possibility to make any assumption on viewpoint repeatability which is, instead, paramount in common datasets oriented to autonomous driving.
The inherent properties of Gaussian Processes, to compensate measurement noise and missing data in the training set (due to holes in the original point clouds), make robust matching possible and opens the possibility of performing robust multi-agent SLAM with different instrument setups in unstructured environments following unconstrained paths.
Furthermore, we highlighted the fact that, although generally performing worse in our specific datasets, visual place recognition is not limited by the availability of depth and therefore the approaches can be fused exploiting multi-modality for more comprehensive relocalization capabilities.
As part of additional future works, we plan to exploit semantic segmentation of known and unknown terrain classes to enhance the understanding and categorization of data within GPGMaps, allowing to improve matching performances in the face of extremely ambiguous structural cues.

\section*{ACKNOWLEDGMENT}
This work was supported by the UA-DAAD 2018 Australia-Germany Joint Research Cooperation Scheme, project COSMA (contract number 57446007) and by the Helmholtz Association, project ARCHES (contract number ZT-0033).

\bibliographystyle{apalike}
\bibliography{bibliography,bibliography_cedric}

\end{document}